\documentclass[10pt,journal,compsoc]{IEEEtran}

\usepackage{ifpdf}

\ifCLASSOPTIONcompsoc
  \usepackage[nocompress]{cite}
\else
  \usepackage{cite}
\fi

\ifCLASSINFOpdf
   \usepackage[pdftex]{graphicx}

\else

   \usepackage[dvips]{graphicx}
\fi

\usepackage{amsmath,amssymb}

\usepackage{algorithmic}
\newsavebox{\ieeealgbox}

\usepackage{array}

\ifCLASSOPTIONcompsoc
  \usepackage[caption=false,font=footnotesize,labelfont=sf,textfont=sf]{subfig}
\else
  \usepackage[caption=false,font=footnotesize]{subfig}
\fi

\usepackage{fixltx2e}
\usepackage{stfloats}

\ifCLASSOPTIONcaptionsoff
  \usepackage[nomarkers]{endfloat}
 \let\MYoriglatexcaption\caption
 \renewcommand{\caption}[2][\relax]{\MYoriglatexcaption[#2]{#2}}
\fi

\newcommand{\removelatexerror}{\let \@latex@error \@gobble}

\usepackage{url}

\usepackage{amsfonts,amsthm,amsmath}
\usepackage{multirow}
\usepackage{graphicx}
\usepackage{algorithm2e}
\RestyleAlgo{ruled}

\newcommand{\comment}[1]{}
\newcommand{\shortcomment}[1]{}
\newcommand{\later}[1]{$\ast\ast\ast$}

\usepackage{amsmath,amsfonts,bm}

\def\Figref#1{Fig.~\ref{#1}}
\def\Tabref#1{Table~\ref{#1}}
\def\Thmref#1{Theorem~\ref{#1}}

\def\Secref#1{Section~\ref{#1}}

\def\eqref#1{equation~\ref{#1}}

\def\1{\bm{1}}

\def\0{{\bm{0}}}
\def\1{{\bm{1}}}

\def\vepsilon{\bm{\epsilon}}

\def\vmu{\bm{\mu}}

\def\vs{{\bm{s}}}

\def\vw{{\bm{w}}}
\def\vx{{\bm{x}}}

\DeclareMathAlphabet{\mathsfit}{\encodingdefault}{\sfdefault}{m}{sl}
\SetMathAlphabet{\mathsfit}{bold}{\encodingdefault}{\sfdefault}{bx}{n}

\DeclareMathOperator{\sign}{sign}

\usepackage{color,soul}
\usepackage{enumitem}
\usepackage{nicefrac}
\usepackage{booktabs}

\renewcommand{\eqref}[1]{(\ref{#1})}

\newtheorem{theorem}{Theorem}
\newtheorem{lemma}{Lemma} 
\def\0{\bm{0}}

\newcommand{\etal}{{et al.}}

\begin{document}
%

\newcommand{\customtitle}{BayesNAM: Leveraging Inconsistency for Reliable Explanations}

\title{\customtitle}


\author{Hoki~Kim,
        Jinseong~Park,
        Yujin~Choi,
        Seungyun~Lee,
        and~Jaewook~Lee
\IEEEcompsocitemizethanks{\IEEEcompsocthanksitem 
Hoki Kim is with the Department of Industrial Security, Chung-Ang University, South Korea. E-mail: hokikim@cau.ac.kr.}
\IEEEcompsocitemizethanks{\IEEEcompsocthanksitem 
Jinseong Park, Yujin Choi, Seungyun Lee, and Jaewook Lee are with the Department of Industrial Engineering, Seoul National University, South Korea. The corresponding author is Jaewook Lee. E-mail: jaewook@snu.ac.kr}
}

\IEEEtitleabstractindextext{%
\begin{abstract}
Neural additive model (NAM) is a recently proposed explainable artificial intelligence (XAI) method that utilizes neural network-based architectures. Given the advantages of neural networks, NAMs provide intuitive explanations for their predictions with high model performance.
In this paper, we analyze a critical yet overlooked phenomenon: NAMs often produce inconsistent explanations, even when using the same architecture and dataset. Traditionally, such inconsistencies have been viewed as issues to be resolved. However, we argue instead that these inconsistencies can provide valuable explanations within the given data model.
Through a simple theoretical framework, we demonstrate that these inconsistencies are not mere artifacts but emerge naturally in datasets with multiple important features. To effectively leverage this information, we introduce a novel framework, Bayesian Neural Additive Model (BayesNAM), which integrates Bayesian neural networks and feature dropout, with theoretical proof demonstrating that feature dropout effectively captures model inconsistencies.
Our experiments demonstrate that BayesNAM effectively reveals potential problems such as insufficient data or structural limitations of the model, providing more reliable explanations and potential remedies.

\end{abstract}

}

\maketitle

\IEEEdisplaynontitleabstractindextext

%
\IEEEpeerreviewmaketitle

\IEEEraisesectionheading{\section{Introduction}\label{sec:introduction}}

\IEEEPARstart{E}{xplainable} artificial intelligence (XAI) has become a significant field of research as machine learning models are increasingly applied in real-world systems including finance and healthcare. To provide insight into the underlying decision-making process behind the predictions made by these models, numerous researchers have developed various techniques to assist human decision-makers.

Recently, Agarwal \etal\cite{agarwal2021neural} proposed a neural additive model (NAM) that utilizes neural networks to achieve both high performance and explainability. NAM is a type of generalized additive model (GAM) that involves the linear or non-linear transformation of each input and yields the final prediction through an additive operation. Previous studies have demonstrated that NAM not only learns complex relationships between inputs and outputs but also provides a high level of explainability based on neural network architectures and training techniques.

\begin{figure}[t!]
    \centering
    \includegraphics[width=0.95\linewidth]{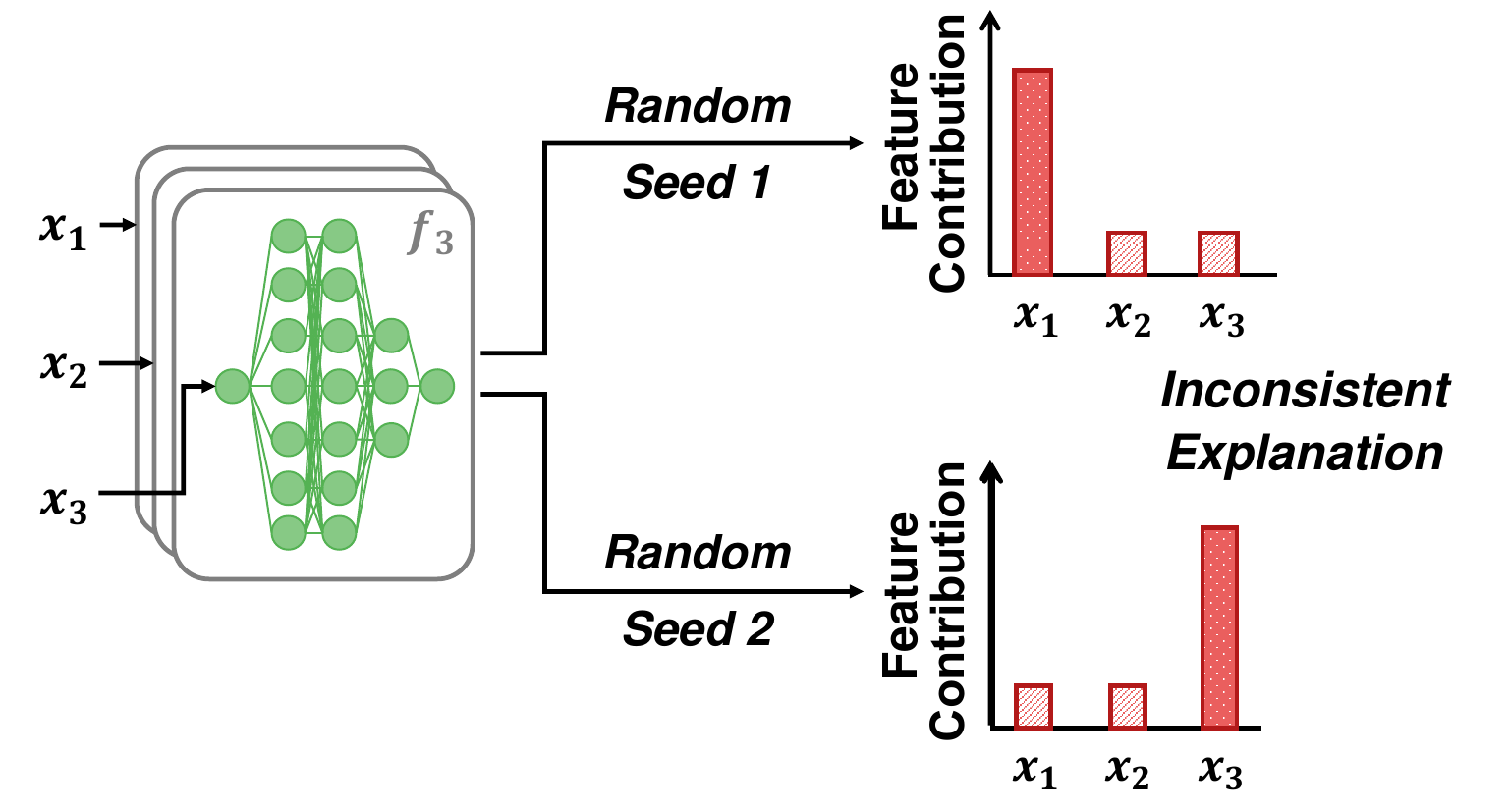} 
    \caption{Inconsistency of NAM, where two independent NAMs trained with the same dataset and architecture output different explanations solely due to different random seeds.}
    \label{fig:intro}
\end{figure}

In this paper, we analyze a critical yet overlooked phenomenon: the inconsistency phenomenon of NAM. \Figref{fig:intro} illustrates this issue, where two independent NAMs, trained on the same dataset and architecture, produce different explanations due solely to variations in random seeds. Such inconsistency has traditionally been viewed as a problem to be solved \cite{radenovic2022neural}. 

However, we argue that these inconsistencies are not merely obstacles but can offer valuable insights to uncover external explanations within the data model.
Through a simple theoretical model, we show that NAMs naturally exhibit the inconsistency phenomenon even when trained on usual datasets that contain multiple important features. Building on this insight, we propose the \textit{Bayesian Neural Additive Model (BayesNAM)}, a novel framework that combines Bayesian neural networks with feature dropout to harness these inconsistencies for more reliable explainability. We also provide theoretical proof that feature dropout effectively leverages inconsistency. Our real-world experiments demonstrate that BayesNAM not only provides more reliable and interpretable explanations but also highlights potential issues in the data model, such as insufficient data and structural limitations within the model.

The main contributions can be summarized as follows:
    \begin{itemize}
    \item We investigate the inconsistency phenomenon of NAMs and analyze this phenomenon through a simple theoretical model.
    \item We propose a new framework BasyesNAM, which utilizes Bayesian neural network and feature dropout. We also establish a theoretical analysis of the efficacy of feature dropout in leveraging inconsistency information.
    \item We empirically demonstrate that BayesNAM is particularly effective in identifying data insufficiencies or structural limitations, offering more reliable explanations and insights for decision-making.
\end{itemize}


\section{Related Work}

\subsection{Neural Additive Model}
As numerous machine learning and deep learning models are black-box, a line of work has attempted to explain the decisions made by a black-box model. We call these methods \textit{post-hoc methods} since they are applied after the model has been trained. 
While post-hoc methods offer some interpretability, recent work \cite{kumar2020problems, rudin2019stop} has argued that these methods can lead to unreliable explanations, which could potentially have detrimental effects on their explainability. 

\begin{figure}[t!]
    \centering
    \includegraphics[width=0.7\linewidth]{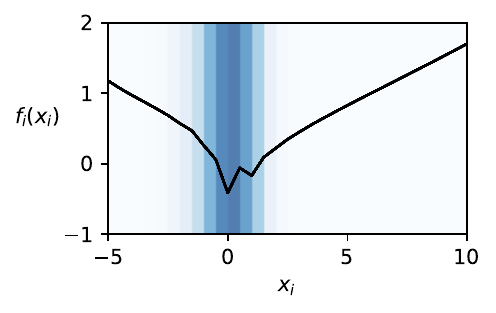} 
    \caption{Example of a mapping function $f_i$ of NAM. Blue regions correspond to regions with high data density. NAM enables us to capture non-linear relationships between inputs and outputs and further provide a clear understanding.}
    \label{fig:nam}
\end{figure}

In contrast to post-hoc methods, \textit{intrinsic methods} aim to develop an inherently explainable model without additional techniques \cite{rudin2019stop}. 
Agarwal \etal\cite{agarwal2021neural} proposed a neural additive model (NAM), which combines a generalized additive model \cite{hastie1990generalized} and neural networks. To be specific, given $d$ features, $x_1, x_2, \cdots, x_d$ and a target $y$, NAM constructs $d$ mapping functions as follows:
\begin{equation}\label{eq:nam}
    y = f_1(x_1) + f_2(x_2) + \cdots + f_d(x_d) + \beta,
\end{equation}
where $\beta$ is a bias term and each mapping function $f_i$. 
In \Figref{fig:nam}, We illustrate an example of NAM.
By utilizing the neural network, NAMs capture the non-linear relationship and achieve high performance while maintaining clarity through a straightforward plot.

Despite their strengths, NAMs frequently exhibit inconsistent explanations even when trained on identical datasets with the same architectures, as illustrated in \Figref{fig:intro}. These inconsistencies can also be observed in the original work \cite{agarwal2021neural}, where the mapping functions produced by different NAMs within an ensemble show substantial variation, despite being trained under the same experimental conditions.

While this inconsistent phenomenon across NAMs can harm its explainability as they are intended to be XAI models, this phenomenon has received limited attention in the literature. To the best of our knowledge, only one study has explicitly addressed this issue. Radenovic et al. \cite{radenovic2022neural} introduced the neural basis model (NBM), which used shared basis functions across features rather than assigning independent mapping functions to each feature. They argued that NBM reduces divergence between models, offering more consistent shape functions compared to NAM, thus mitigating the inconsistency problem.

In contrast, this paper presents a novel view on the inconsistency phenomenon. Rather than treating it as a problem to be solved, we argue that these inconsistencies provide valuable information about the data model. 

\subsection{Bayesian Neural Network}
Although the use of a single model is a fundamental approach, numerous studies have found that a point-estimation is often vulnerable to overfitting and high variance due to its limited representation \cite{dietterich2000ensemble}. 
To overcome this limitation, Bayesian neural network estimates the model distribution instead of calculating a fixed model. Given the data $(\vx, y) \sim \mathcal{D}$ and the prior $p(\vw)$, we aim to approximate the posterior $p(\vw|\vx, y)$. Specifically, rather than using a fixed weight vector $\vw_i$, it aims to find a distribution of weight vectors $\mathcal{N}(\vmu_i, \texttt{diag}(\vs_i)^2)$ and learn the mean vector $\vmu_i$ and the standard deviation vector $\vs_i$.

Since the distribution $p(\vx, y)$  is generally intractable, several methods have been developed to approximate the posterior, including Markov Chain Monte Carlo (MCMC) \cite{welling2011bayesian} and variational inference approaches \cite{graves2011practical, blundell2015weight}. While MCMC methods can provide more accurate estimates, their high computational cost \cite{li2016preconditioned} has led to the use of variational inference methods across diverse domains \cite{liu2018adv,lee2022graddiv}.

During optimization in variational inference methods, an weight vector $\vw_i=\vmu_i + \vs_i\odot \vepsilon$ is sampled for each forward step where $\vepsilon \sim \mathcal{N}(\mathbf{0}, \mathbf{I})$. The prior distribution can be simply chosen as the isometric Gaussian prior $\mathcal{N}(\mathbf{0}, s_0^2\mathbf{I})$ where $s_0$ is a predefined standard deviation to explicitly calculate the KL-divergence \cite{liu2018adv}. 

A promising direction in the field of Bayesian neural networks is their integration with other domains to enhance model explainability. Bayesian neural networks provide weight distributions that enable the identification of high-density regions or confidence intervals, which can be used for uncertainty estimation. Researchers and practitioners in several domains that require reliable explanations, such as medicine \cite{singh2021uncertainty} and finance \cite{jang2019generative}, have also explored the utilization of Bayesian models to measure the confidence of prediction for trustworthy decision-making. 

\section{Methodology} \label{sec:theo}
In \Secref{subsec:inconsistency}, we first investigate the inconsistency phenomenon of NAMs with a simple theoretical model. Our empirical findings show that this inconsistency can easily occur, even when datasets contain more than one important feature. Subsequently, in \Secref{subsec:bayesnam}, we propose a new framework called BayesNAM, which combines Bayesian neural network with feature dropout, to leverage the inconsistency information as a source of valuable indicator. This framework is supported by a theoretical analysis demonstrating the effectiveness of feature dropout in capturing diverse explanations. Finally, we provide a detailed explanation of the proposed framework.

\subsection{Rethinking Inconsistency of Neural Additive Model}
\label{subsec:inconsistency}

\begin{figure}[t!]
    \centering
    \subfloat[Random Seed 1 \label{fig:lam0_1}]{%
       \includegraphics[height=0.9\linewidth]{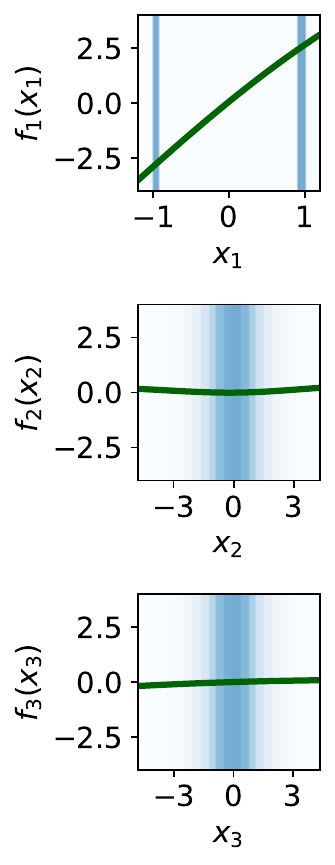}\quad\quad}
    \subfloat[Random Seed 2\label{fig:lam0_2}]{%
       \includegraphics[height=0.9\linewidth]{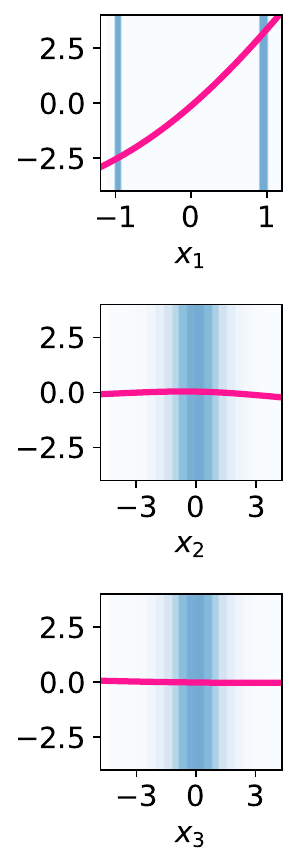}}
    \caption{(Case-I. $\lambda=0$) Mapping functions of two NAMs trained with different random seeds show similar shapes. Blue regions correspond to regions with high data density.}
    \label{fig:lam0}
\end{figure}

We begin the analysis by identifying and investigating the inconsistent explanations of NAM. To this end, we construct a simple theoretical model.
Here, we consider a binary classification task where the target $y$ can have a value in $\{-1, 1\}$. Inspired by \cite{tsipras2018robustness}, we construct the input-target pairs $(\vx, y)=(x_1, x_2, \cdots, x_d, y)$ from a distribution $\mathcal{D}$ as follows:
\begin{align} \label{eq:toy}
    x_1 &= \begin{cases}
          +y \text{ with probability } p \\
          -y \text{ with probability } 1-p
        \end{cases},
    \\
    x_2&, \cdots, x_d \stackrel{i.i.d.}{\sim} \mathcal{N}(\lambda y, \sigma^2),
\end{align}
where $x_2, \cdots, x_d$ are independently and identically sampled from a normal distribution $\mathcal{N}$ with the mean $\lambda y$ and the standard deviation $\sigma^2$ for positive $\lambda$ and $\sigma$. It is important to note that the features $x_2, \cdots, x_d$ are uncorrelated, as they are drawn independently and identically distributed. By adjusting the values of $p$ and $\lambda$, we can control the significance of $x_1$ and $x_2, \cdots, x_d$ in predicting $y$, as stated in the following lemma:
\begin{lemma} (Derived from \cite{tsipras2018robustness}) \label{lem:lemma1}
Consider a linear classifier $h$,
\begin{align}\label{eq:h}
    h(x_2, \cdots, x_d) &= \sign(w_2 x_2 + w_3 x_3 + \cdots + w_d x_d).
\end{align}
Then, even a natural linear classifier, $h(\cdot)$ with $w_i= \frac{1}{d-1}$, can easily achieve a higher classification accuracy than $p$, which is a natural accuracy of the model that only uses $x_1$, if the following statement is satisfied:
\begin{align} \label{eq:aa}
    \Phi_{X \sim \mathcal{N}\left(0, \sigma^2/(d-1)\right)}(\lambda) > p,
\end{align}
where $\Phi_X (\cdot)$ is the cumulative distribution function of $X$. (Detailed proof is presented in Appendix)

\end{lemma}
Let $p$ be a sufficiently large positive number. When $\lambda=0$, only $x_1$ is useful to predict $y$ and other features $x_2, \cdots, x_d$ are not correlated to $y$. 
As $\lambda$ increases, $x_2, \cdots, x_d$ become correlated to $y$. By Lemma \ref{lem:lemma1}, if \eqref{eq:aa} is satisfied, a model that only considers $x_2, \cdots, x_d$ can achieve a higher classification accuracy than $p$.
In summary, if $\lambda=0$, $x_1$ would be the only feature with a high importance in predicting $y$, while $x_2, \cdots, x_d$ is enough to have a significant performance in predicting $y$ for a large $\lambda>0$.

Now, we consider the following two cases with d=3:
\begin{itemize}
\item\textbf{Case-I.} \textbf{Single important feature exists} $(\lambda=0)$.
In this case, only $x_1$ is effective in predicting $y$, while $x_2$ and $x_3$ are not useful.

\item\textbf{Case-II.} \textbf{Multiple important features exist} $(\lambda=3)$.
In this case, all the features, $x_1$, $x_2$, and $x_3$ are highly correlated with $y$. The model uses $x_2$ and $x_3$ can perform better than the model sorely depends on $x_1$ since $\Phi_{Z}(\lambda=3)=0.999$.
\end{itemize}

Given this theoretical model, we generated two sets of data containing 50,000 training examples and 10,000 test examples and trained two different NAMs on each dataset for different random seeds. For simplicity, we fixed the feature dimension to $d=3$, the probability $p=0.95$, and $\sigma^2=d-1$, resulting \eqref{eq:aa} becomes $\Phi_{Z}(\lambda)>p$, where $Z$ is drawn from the standard normal distribution $\mathcal{N}(0, 1)$. For each mapping function $f_i$ of NAM, we constructed a simple neural network with two linear layers containing 10 hidden neurons and used ReLU as an activation function. The models are trained by SGD with a learning rate of 0.01. One epoch was sufficient to achieve high training accuracy.

\begin{figure}[t!]
    \centering
    \subfloat[Random Seed 1 \label{fig:lam3_1}]{%
       \includegraphics[height=0.9\linewidth]{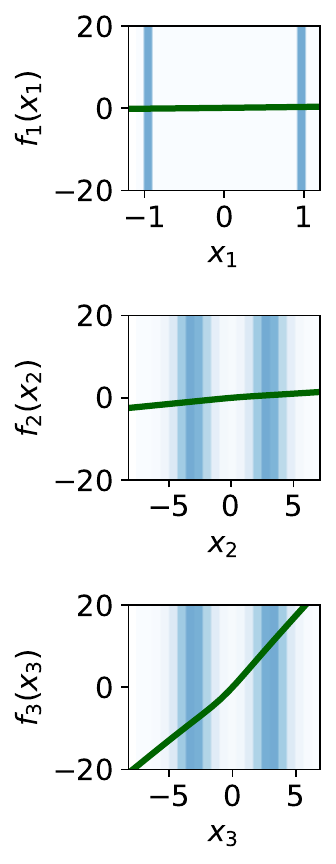}\quad\quad}
    \subfloat[Random Seed 2\label{fig:lam3_2}]{%
       \includegraphics[height=0.9\linewidth]{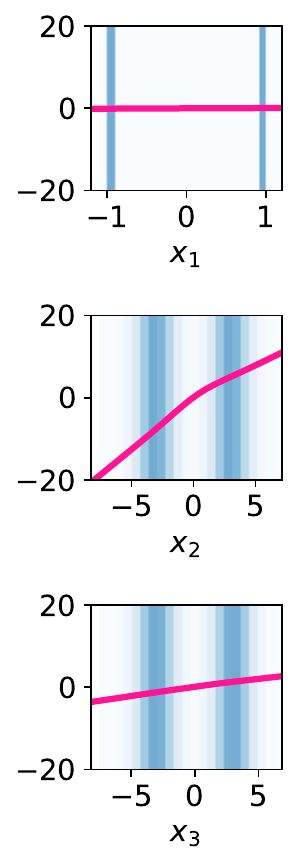}}
    \caption{(Case-II. $\lambda=3$) Mapping functions of two NAMs trained with different random seeds are extremely different. This yields inconsistent feature contribution in \Figref{fig:inconsistent}.}
    \label{fig:lam3}
\end{figure}

\Figref{fig:lam0} (Case-I) and \Figref{fig:lam3} (Case-II) illustrate the mapping functions of trained NAMs for each case. Specifically, for Case-I, we observed that the two NAM models trained with different random seeds exhibited similar test accuracy and explanations (94.9\% and 95.0\%, respectively). As shown in \Figref{fig:lam0}, the mapping functions for each $x_i$ have similar shapes, with $f_1$ being the only increasing one and the others being almost constant. Therefore, in this case, NAM successfully captures the true importance of features and provides reliable explanations.

\begin{figure}[t!]
    \centering
    \includegraphics[width=0.7\linewidth]{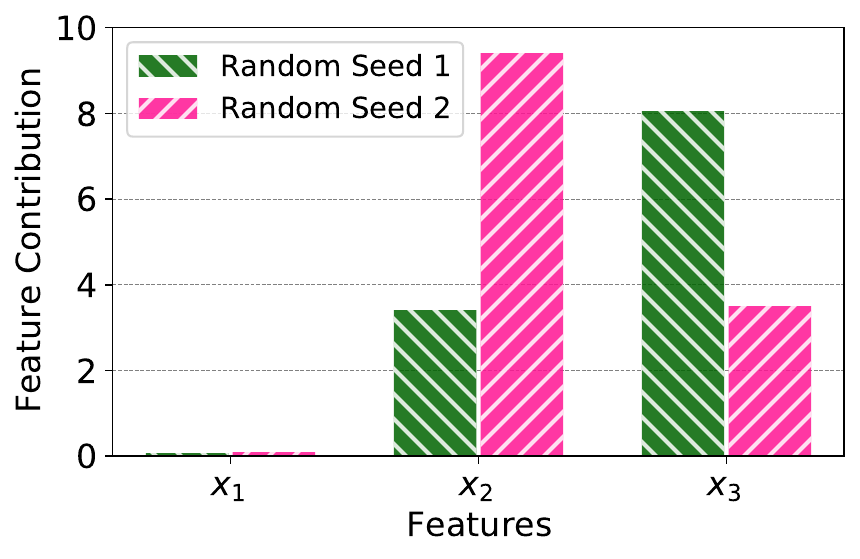}  
    \caption{Corresponding feature contributions of a sample ${\vx}=[-1, 3, 3]$ with $y=1$ for NAMs in \Figref{fig:lam3}. This inconsistent explanation corresponds to \Figref{fig:intro}.}
    \label{fig:inconsistent}
\end{figure}

In contrast, for Case-II (\Figref{fig:lam3}), the mapping functions $f_i$ of the trained NAMs have extremely different shapes. Although both NAMs achieve a test accuracy exceeding 99.99\%, $f_3$ is much steeper than $f_2$ for the first random seed, whereas the relationship is reversed for the second random seed, and vice versa.

Such inconsistency results in inconsistent feature contribution. Figure \ref{fig:inconsistent} shows the feature contribution of a sample $\vx=[x_1,x_2,x_3]$ $=[-1, 3, 3]$. Following \cite{agarwal2021neural}, we calculate the feature contribution by subtracting the average value of a mapping function across the entire training dataset. Although we use the same example, the feature contribution calculated by NAM with random seed 1 implies that $x_2$ appears to be more important than $x_3$, while NAM with random seed 2 outputs the opposite result that $x_3$ appears to be more significant than $x_2$.
In summary, NAMs can produce inconsistent explanations when multiple important features are present, a common condition in real-world datasets. Indeed, as discussed later in Figures \ref{fig:COMPAS_juv} and \ref{fig:ca}, this inconsistency is readily observable in widely-used datasets.

At first glance, the observed inconsistency may appear problematic; however, both explanations are not inherently incorrect. Specifically, given the theoretical model, both $x_2$ and $x_3$ are important features under Case-II, as using only one of them can achieve high performance. Therefore, the distinct mapping functions demonstrate that the different perspectives of trained models and each explanation is a valid interpretation of the data model, where relying solely on either $x_2$ or $x_3$ is sufficient for high performance.

\begin{figure}[t!]
    \centering
    \subfloat[Learning Rate \label{fig:bs_new}]{%
       \includegraphics[width=0.8\linewidth]{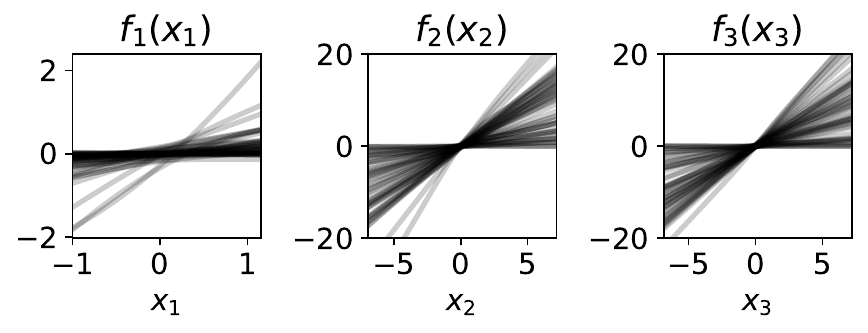}}
       \hfill
    \subfloat[Batch size\label{fig:lr_new}]{%
       \includegraphics[width=0.8\linewidth]{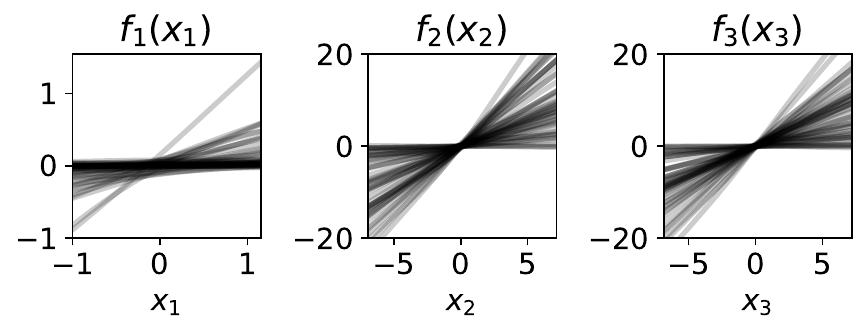}}
    \caption{Mapping functions of NAMs trained with different learning rates and batch sizes. Given the fact that all models achieve more than 99\% test accuracy, this inconsistency tells us that high-performing models can have diverse perspectives for the given data model.}
    \label{fig:other}
\end{figure}

In \Figref{fig:other}, we vary the learning rates ($\eta$) and batch sizes ($B$) during training within the same theoretical model. We linearly increase the learning rate $\eta$ from 0.005 to 0.01, and the batch size $B$ from 5 to 50. In total, we trained 50 models for each experiment, where each NAM exhibits inconsistent mapping functions. 
However, it is important to note that all models achieved over 99\% test accuracy on the dataset. This indicates that the diverse explanations are not incorrect; rather, they offer valuable external insights into the existence of diverse perspectives among high-performing models, complementing the internal explanations of individual models. Therefore, we posit that \textbf{inconsistency can be a useful indicator of potential external explanatory factors.} Based on these findings, we propose a new framework to leverage inconsistency and provide additional explanations within the data model.

\subsection{Bayesian Neural Additive Model} \label{subsec:bayesnam}
In the previous subsection, we explored the inconsistency phenomenon in NAMs and suggested that rather than being a flaw, this inconsistency can serve as a valuable source of additional information, shedding light on underlying external explanations in the data model. In this section, we introduce BayesNAM, a novel framework designed to leverage this inconsistency. BayesNAM incorporates two key approaches: (1) a modeling approach based on Bayesian structure and (2) an optimization approach utilizing feature dropout. 
Each of these approaches will be detailed in the following paragraphs.

\textbf{1) Modeling Approach: Bayesian Structure for Inconsistency Exploration.}
A naive approach to exploring possible inconsistencies in NAMs is by training multiple independent models. Indeed, Agarwal \etal\cite{agarwal2021neural} trained several NAMs and visualized the learned shape functions $f_k(x_k)$. However, this requires training $n$ independent models, leading to a computational burden proportional to $n$, making it impractical for large-scale applications.

To address this limitation, we propose using Bayesian neural networks \cite{graves2011practical, blundell2015weight}, which inherently allow for efficient exploration of model uncertainty without the need to train multiple independent models. Under variational inference and Bayes by Backprop \cite{graves2011practical, blundell2015weight}, Bayesian neural networks rather train the mean parameter $\vmu_i$ and the standard deviation parameter $\vs_i$ instead of an weight vector $\vw_i$. Then, during the training and inference phase, it samples a weight $\vw_i=\vmu_i + \vs_i\odot \vepsilon$ for a random vector $\vepsilon$ from a predefined distribution. 
Following prior works \cite{liu2018adv,lee2022graddiv}, we adopt the reparameterization trick \cite{kingma2015variational} for efficient training. This results in the following training objective.
\begin{align}\label{eq:bayes}
    &\underset{\vmu_i,\vs_i}{\texttt{min}} \mathcal{L}\left(\sum_{i=1}^d f_i(x_i|\vmu_i,\vs_i)+\beta, y\right) + \sum_{i=1}^d\texttt{KL}\left(q_{\vmu_i,\vs_i}(\vw_i)\lVert p(\vw_i)\right),
\end{align}
where $\mathcal{L}(\cdot)$ is a given loss function and $\texttt{KL}(\cdot \lVert\cdot)$ is the KL-divergence. 
For further details, we refer the readers to \cite{blundell2015weight}.

\begin{figure*}[t]
\centering
    \includegraphics[width=0.8\linewidth]{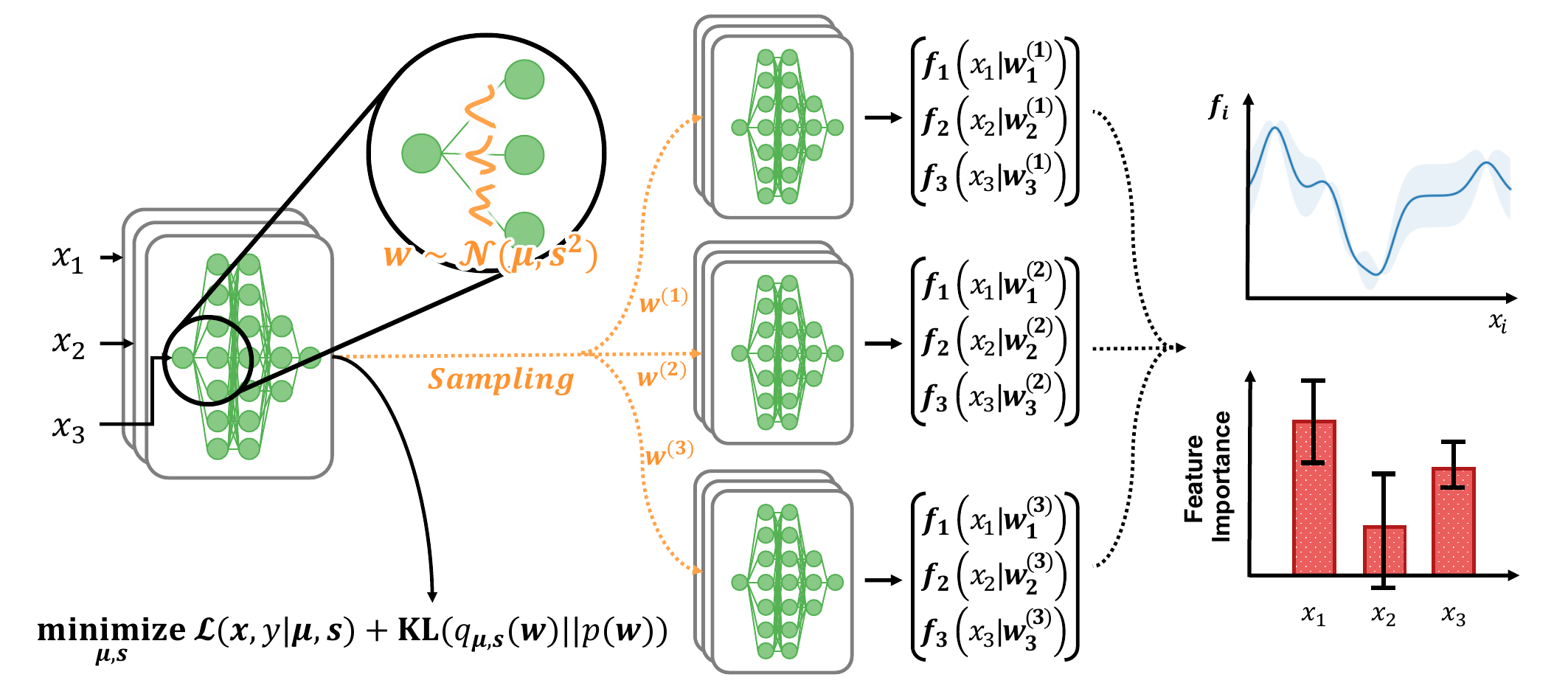} 
  \caption{Structural framework of BayesNAM. BayesNAM trains the distribution of parameters through updating $\vmu$ and $\vs$. During the inference phase, based on the weights $\vw^{(j)}$ drawn from the trained distribution, it can provide rich explanations for its prediction by leveraging inconsistency, such as the confidence interval (upper) of feature contribution (lower).}
  \label{fig:vol}
\end{figure*}

In \Figref{fig:vol}, we present the structural framework that integrates a Bayesian neural network with a neural additive model. For each sampled weight $\vw^{(j)}$, we compute the corresponding predictions $f_i(x_i | \vw^{(j)}_i)$. This sampling approach enables the model to efficiently explore a diverse range of model spaces without needing to train multiple models. By incorporating a Bayesian neural network, the model provides high-density regions of the mapping functions and provides confidence intervals for feature contributions, offering richer interpretability.

\begin{figure}
    \centering
    \subfloat[Naive Bayesian \label{fig:bayes}]{%
       \includegraphics[height=0.196\linewidth]{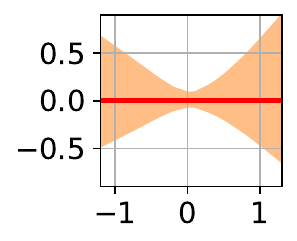}
       \includegraphics[height=0.192\linewidth]{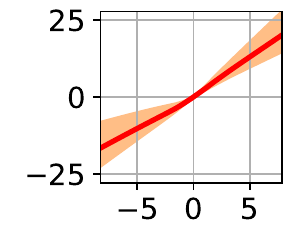}\quad}
    \subfloat[w/ Feature Dropout \label{fig:bayes_fd}]{%
       \includegraphics[height=0.196\linewidth]{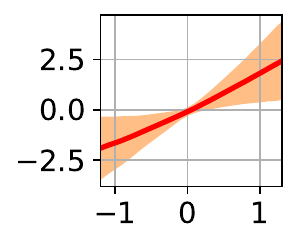}\includegraphics[height=0.192\linewidth]{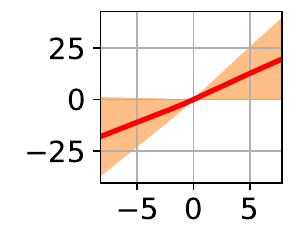}\quad\quad}
  \caption{Effectiveness of feature dropout. The same setting is used as in \Figref{fig:lam3}. Each plot shows the mapping functions of $x_2$ (left) and $x_3$ (right). Both models use the structural framework depicted in \Figref{fig:vol}. Without feature dropout (\Figref{fig:bayes}), the model tends to focus on a single feature, similar to training a single NAM. In contrast, feature dropout (\Figref{fig:bayes_fd}) enables the model to explore diverse explanations.}
  \label{fig:fd}  
\end{figure}

\textbf{2) Optimization Approach: Feature Dropout for Encouraging Diverse Explanations.}
Although Bayesian neural networks provide an efficient mechanism for exploration, they do not inherently guarantee exploring diverse explanations. Indeed, as shown in \Figref{fig:bayes}, naive Bayesian neural network alone tends to focus on a single feature, similar to training a single NAM, rather than adequately exploring diverse explanations. As previously noted in related works \cite{liu2018adv, lee2022graddiv}, we observe that increasing the standard deviation hyper-parameter $s_0$ within Bayesian neural network tends to degrade model performance and fails to address this issue effectively. Therefore, given the presence of diverse valid explanations shown in Figures \ref{fig:lam3} and \ref{fig:inconsistent}, it is evident that a method is needed to encourage the exploration of diverse explanations.

As a potential solution, we propose the use of feature dropout during optimization. Feature dropout, initially introduced by Agarwal \etal\cite{agarwal2021neural}, extends traditional dropout by selectively omitting individual feature networks during training. The hyperparameter $\tau$ determines the probability of dropping each feature. While the original work focused on improving model performance with feature dropout, we here provide a theoretical analysis showing that feature dropout implicitly encourages diverse explanations, preventing over-relying on any single feature.

Given the theoretical model in \Secref{subsec:inconsistency}, we establish the following theorem.

\begin{theorem} \textbf{(Feature Dropout Implicitly Encourages Exploring Diverse Explanations)} \label{thm:feature_dropout}
Given the dataset $(\vx, y)$ in \eqref{eq:toy}, the linear classifier $h$ in \eqref{eq:h}, and the feature dropout rate $\tau$, without loss of generality, the maximal training accuracy of $h$ that only uses $k$ features becomes
\begin{equation}
    \mathcal{P}(k, \tau) = \mathbb{P}_{x_i \sim \mathcal{N}(\lambda y, \sigma^2), u_i \sim \mathcal{B}(1-\tau)}\left[\frac{y}{k}\sum_{i=2}^{k+1} u_i x_i > 0 \right].
\end{equation}
Then, for $k\geq 3$ and $\tau\in[0, \frac{1}{2}]$, the gap $\Delta\mathcal{P} := \mathcal{P}(k, \tau) - \mathcal{P}(1, \tau)$ is always positive and increases as $\tau$ increases. Thus, the model leverages multiple features to achieve high performance, implicitly encouraging the exploration of diverse explanations.
\begin{proof} [Sketch of proof] 
Let $q_j=\Phi_{Z}(\lambda\sqrt{j}/\sigma)$.  Then, $\mathcal{P}(k, \tau)$ can be formalized as follows:
\begin{equation*}
    \mathcal{P}(k, \tau) = \sum^{k}_{j=1} q_j \binom kj (1-\tau)^j \tau^{k-j}.
\end{equation*}
To show that $\Delta \mathcal{P}(k,\tau)>0$ and $\frac{\partial}{\partial \tau}\Delta \mathcal{P}(k,\tau)>0$ for $k \geq 3$ and 8$\tau\in[0, \frac{3}{2}]$, we use mathematical induction.

Applying Pascal's identity and strong induction, we find
\begin{align*}
\frac{\partial}{\partial \tau}\Delta &\mathcal{P}(k + 1,\tau) =
\frac{\partial}{\partial \tau}\Delta \mathcal{P}(3,\tau) + q_1 \tau^2 (3- (k+1)\tau^{k-2})\\& + (q_2-q_1)\tau(1-\tau)(6-(k+1)k\tau^{k-2}) +3 (q_3-q_2)(1-\tau)^2 \\&- \sum^{k}_{j=2} (j+1)(q_{j+1}-q_j)\binom {k+1} {j+1} (1-\tau)^j \tau^{k-j}
\end{align*}

We now consider two cases for \(\tau\): (1) \(\tau \in \left[0, \frac{3}{8}\right]\) and (2) \(\tau \in \left(\frac{3}{8}, \frac{1}{2}\right]\). In each case, we prove that \(\frac{\partial}{\partial \tau}\Delta\mathcal{P}(k + 1, \tau) > 0\) by finding the value of \(\lambda / \sigma\) that minimizes each term. With these lower bounds, we can conclude that the overall expression is positive. 
\textit{(Detailed proof is presented in Appendix)}
\end{proof}
\end{theorem}

\begin{figure}[t!]
    \centering
    \subfloat[$\lambda=3$ \label{fig:proof_mu3}]{%
       \includegraphics[width=0.49\linewidth]{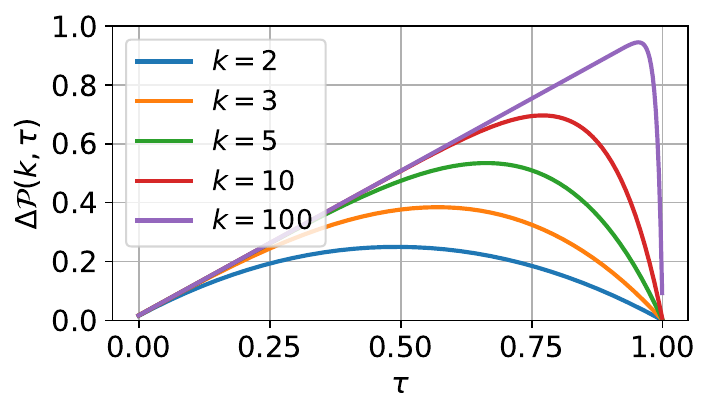}\quad}
    \subfloat[$\lambda=0.01$ \label{fig:proof_mu0}]{%
       \includegraphics[width=0.49\linewidth]{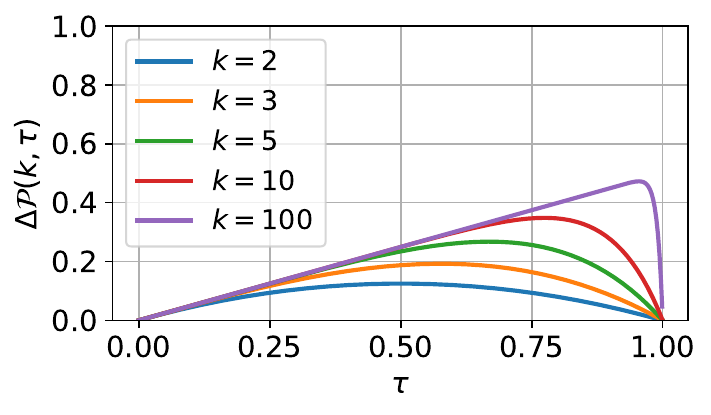}}
    \caption{Empirical verification of \Thmref{thm:feature_dropout}. As $\tau$ increases in range of $[0, \frac{1}{2}]$, $\Delta\mathcal{P}$ increases as well. Moreover, as $k$ increases, the acceptable range of $\tau$ in  \Thmref{thm:feature_dropout} expands.}
    \label{fig:emp_proof}
\end{figure}

\Figref{fig:emp_proof} empirically verifies the general acceptance of \Thmref{thm:feature_dropout}. We plot $\Delta \mathcal{P}(k, \tau)$ with varying $k$. Other settings are same as the Case-II in \Secref{subsec:inconsistency}. The increasing trend of $\Delta \mathcal{P}(k, \tau)$ for $\tau \in [0, \frac{1}{2}]$ aligns with the implications of \Thmref{thm:feature_dropout}. Furthermore, the acceptable range of $\tau$ in  \Thmref{thm:feature_dropout} expands as $k$ increases. When $k=100$, $\Delta \mathcal{P}(k, \tau)$ is increasing for $\tau \in [0, 0.9]$. Even for $k = 2$, we observe that $\Delta \mathcal{P}(k, \tau)$ increases until $\tau = 0.4$.  
Since \Thmref{thm:feature_dropout} holds regardless of the values of $\lambda$, we observe similar results for a very small value of $\lambda = 0.01$.

In summary, we theoretically and empirically verify that feature dropout encourages the model to explore diverse explanations by using multiple features in the dataset. As shown in \Figref{fig:bayes_fd}, incorporating feature dropout enables the model to explore explanations across a range of features. Therefore, we introduce the framework that combines Bayesian neural networks with feature dropout as Bayesian Neural Additive Model (BayesNAM).

\section{Experiments}\label{sec:exp}

In this section, we present empirical findings comparing the performance of our proposed framework against traditional models, such as Logistic/Linear Regression, Classification and Regression Trees (CART), and Gradient Boosted Trees (XGBoost) \cite{chen2016xgboost}, as well as recent explainable models including the Explainable Boosting Machine (EBM) \cite{nori2019interpretml}, NAM, NAM with an ensemble method (NAM+Ens), and our proposed BayesNAM. For Logistic/Linear Regression, CART, XGBoost, and EBM, we conducted a grid search for hyperparameter tuning, following the settings outlined in \cite{agarwal2021neural}. We found that using ResNet blocks—comprising two group convolution layers with BatchNorm and ReLU activation—yields better performance for NAM and BayesNAM compared to the ExU units or ReLU-$n$ suggested in \cite{agarwal2021neural}. For NAM+Ens, we trained five independent NAMs, and both NAM+Ens and BayesNAM utilized soft voting for model aggregation during evaluation. Detailed settings are provided in the Appendix.

We evaluated all models on five different datasets: Credit Fraud \cite{dal2015adaptive}, FICO \cite{fico2018fico}, and COMPAS \cite{compas2016compas} for classification tasks, and California Housing (CA Housing) \cite{pace1997sparse} and Boston \cite{harrison1978hedonic} for regression tasks. 
As shown in \Tabref{tab:performance}, BayesNAM demonstrates comparable performance to other benchmarks across datasets, with particularly strong results in classification tasks such as COMPAS, Credit Fraud, and FICO. For regression tasks, BayesNAM tends to be less accurate, which we discuss further in the Appendix.

\begin{table}[t]
\caption{Performance comparison between models on 5 different random seeds. Higher AUC is better ($\uparrow$) and lower RMSE is better ($\downarrow$).}
\label{tab:performance}
\centering
\resizebox{1\linewidth}{!}{%
\begin{tabular}{|c|c|c|c|c|c|}
\hline
Model              & \begin{tabular}[c]{@{}c@{}}COMPAS\\ (AUC$\uparrow$)\end{tabular} & \begin{tabular}[c]{@{}c@{}}Credit\\ (AUC$\uparrow$)\end{tabular} & \begin{tabular}[c]{@{}c@{}}FICO\\ (AUC$\uparrow$)\end{tabular} & \begin{tabular}[c]{@{}c@{}}Boston   \\ (RMSE$\downarrow$)\end{tabular} & \begin{tabular}[c]{@{}c@{}}CA   Housing\\ (RMSE$\downarrow$)\end{tabular} \\ \hline\hline
Log./Lin.   Reg.  & 0.699$\pm$0.005 & 0.977$\pm$0.004 & 0.706$\pm$0.005 & 5.517$\pm$0.009 & 0.731$\pm$0.010 \\ 
CART            & 0.776$\pm$0.005 & 0.956$\pm$0.005 & 0.784$\pm$0.002 & 4.133$\pm$0.004 & 0.712$\pm$0.007 \\ 
XGBoost            & 0.743$\pm$0.012 & 0.980$\pm$0.005 & 0.795$\pm$0.001 & 3.155$\pm$0.009 & 0.531$\pm$0.011 \\ \hline\hline
EBM                & 0.764$\pm$0.009   & 0.978$\pm$0.007   & 0.793$\pm$0.005   & 3.301$\pm$0.005   & 0.558$\pm$0.012   \\
NAM                & 0.769$\pm$0.011 & 0.989$\pm$0.007 & 0.804$\pm$0.003 & 3.567$\pm$0.012 & 0.556$\pm$0.009 \\ 
NAM+Ens               & 0.771$\pm$0.005 & 0.990$\pm$0.004 & 0.804$\pm$0.002 & 3.555$\pm$0.006 & 0.554$\pm$0.003 \\ 
BayesNAM          & 0.784$\pm$0.009   & 0.991$\pm$0.003   & 0.804$\pm$0.001   & 3.620$\pm$0.011   & 0.556$\pm$0.007   \\ 
\hline
\end{tabular}%
}
\end{table}

\subsection{Identifying Data Inefficiency}
\begin{figure*}
    \centering
    \begin{minipage}[c]{0.79\linewidth}
        \centering
    {%
       \includegraphics[width=\linewidth]{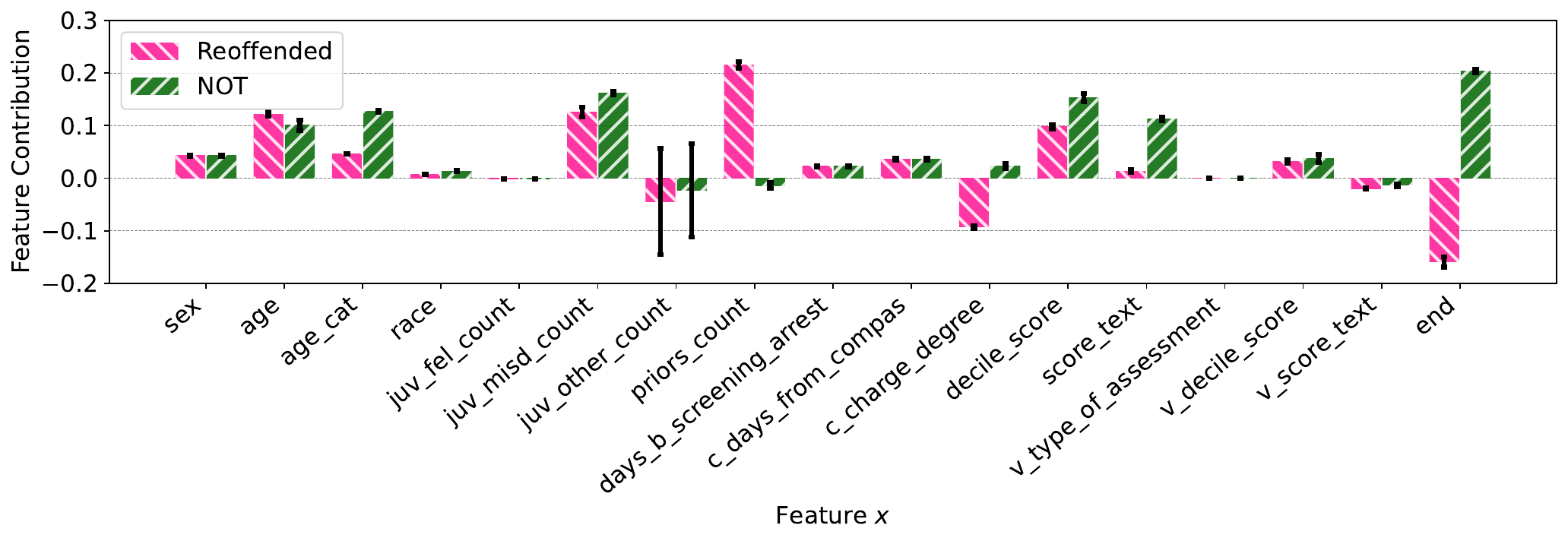}}
    \end{minipage}\hfill
    \begin{minipage}[c]{0.2\linewidth}
        \centering
{%
       \includegraphics[width=\linewidth]{figs/COMPAS_juv\_other\_count2.pdf}}
    \end{minipage}
    \caption{Empirical results on COMPAS. (Left) Feature contribution of two randomly drawn offenders obtained from BayesNAM. Error bars correspond to the standard deviation of feature contribution. \texttt{juv\_other\_count} shows an extremely high variance. (Right) Mapping function of \texttt{juv\_other\_count}. Gray lines represent the mapping functions from five different NAMs, while the red line and orange shaded area indicate the average mapping function and its two-sigma interval for BayesNAM, respectively.  The mapping functions begin to diverge, i.e., inconsistency occurs, when \texttt{juv\_other\_count} $\geq$ 4.}
    \label{fig:COMPAS_juv}
\end{figure*}

The capability of BayesNAM to explore diverse explanations further allows us to obtain confidence information of feature contributions. In the left plot of Figure \ref{fig:COMPAS_juv}, we plot the feature contributions of two randomly drawn offenders from each target value, `reoffended' ($y=1$) or `not' ($y=0$).

\begin{figure}[t!]
    \centering
    \subfloat[Data distribution within \texttt{juv\_other\_count} \label{fig:Hist_COMPAS}]{%
       \includegraphics[width=0.75\linewidth]{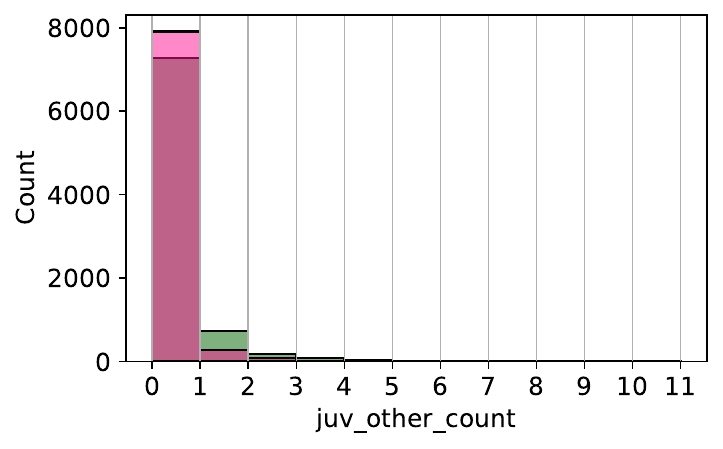}\quad\quad\quad}
       \hfill
    \subfloat[Class imbalance in \texttt{juv\_other\_count} $\geq$ 9. \label{fig:Hist_COMPAS_ratio}]{%
       \includegraphics[width=0.8\linewidth]{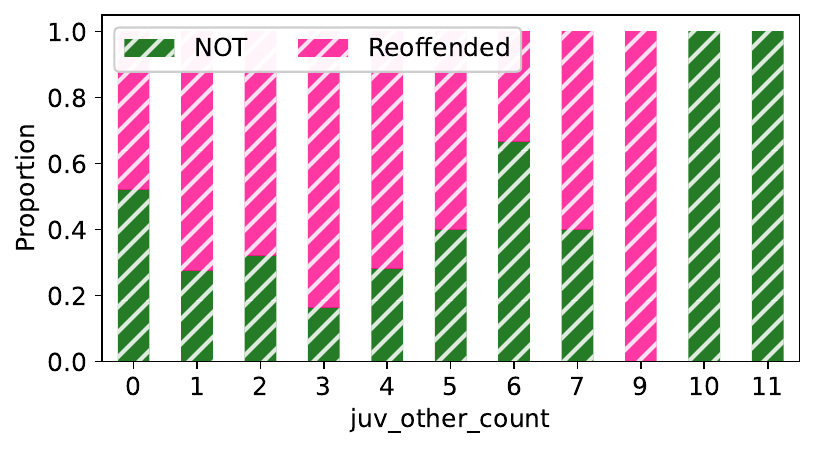}\quad\quad}
       \hfill
    \caption{Empirical findings on COMPAS. (Top) The high variance area corresponds to a lack of data for \texttt{juv\_other\_count} $\geq 4$. (Bottom) This range also shows skewed proportions, particularly for \texttt{juv\_other\_count} $\geq 9$, where all labels are either 'reoffended' or 'not.'}
    \label{fig:hist}
\end{figure}

Among the features, \texttt{juv\_other\_count} (which represents the number of non-felony juvenile offenses a person has been convicted of) exhibits high variance in its contributions. This high variance indicates that, with a single NAM, the contribution of \texttt{juv\_other\_count} can appear either extremely negative or positive, potentially leading to misinterpretation. BayesNAM reveals substantial variability among models, suggesting that \texttt{juv\_other\_count} can have both positive and negative effects on predictions within certain ranges.

What can we infer from this high variation? In the right plot of Figure \ref{fig:COMPAS_juv}, we analyze the mapping function of \texttt{juv\_other\_count}. NAMs (gray) show inconsistent explanations for \texttt{juv\_other\_count}. The two-sigma interval of mapping functions of BayesNAM (orange) also starts to diverge significantly when \texttt{juv\_other\_count} $\geq 4$, indicating increased inconsistency in this range. 
As shown in Figure \ref{fig:Hist_COMPAS}, a data range where \texttt{juv\_other\_count} $\geq 4$ indicates a lack of sufficient data. Moreover, this range also shows skewed proportions, especially for \texttt{juv\_other\_count} $\geq 9$, where all labels are either 'reoffended' or 'not.' In summary, we verify that the high inconsistency highlights the need for caution when interpreting examples involving the feature, and suggests potential issues such as a lack of data.

In addition to identifying data insufficiencies, our model can also be used for feature selection. Features with high absolute contributions and small standard deviations, such as \texttt{priors\_count} (which represents the total number of prior offenses a person has been convicted of), consistently demonstrate significant impact across different models.

\subsection{Capturing Structural Limitation}

\begin{figure}[t!]
    \centering
    \subfloat[Model prediction \label{fig:CA_Longitude}]{%
       \includegraphics[width=0.6\linewidth]{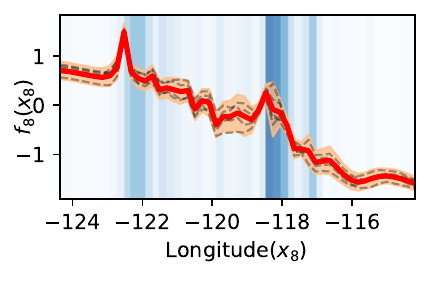}}
       \hfill
    \subfloat[Price visualization \label{fig:CA_map}]{%
       \quad\includegraphics[width=0.65\linewidth]{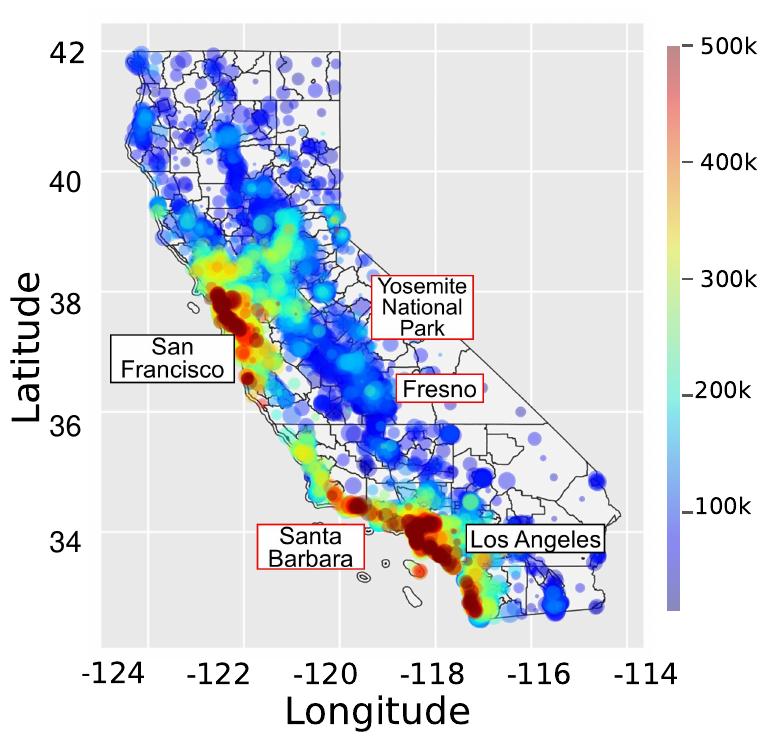}}
    \caption{Empirical findings on CA. (Top) Mapping functions for \texttt{Longitude}, similar to the right plot of \Figref{fig:COMPAS_juv}. (Bottom) Housing prices are represented with colors in thousand dollars.}
    \label{fig:ca}
\end{figure}

\begin{figure*}
    \centering
    \begin{minipage}[c]{0.2\linewidth}
        \centering
        \includegraphics[width=\linewidth]{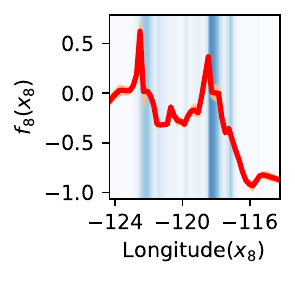} \label{fig:CANEW_Longitude}
    \end{minipage}
    \begin{minipage}[c]{0.7\linewidth}
        \centering
        \includegraphics[width=\linewidth]{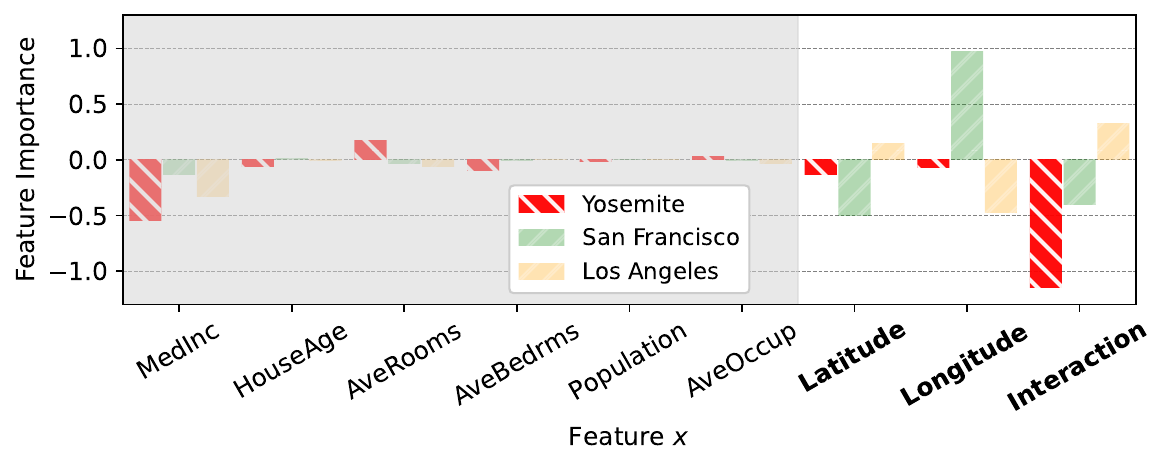}
        \label{fig:Barchart_HousingNew}
    \end{minipage}
    \caption{High variance observed in \Figref{fig:CA_Longitude} suggests the potential failure of model assumption. (Left) Based on our analysis in \Figref{fig:ca}, we construct and train a new model that contains the interaction term between Latitude and Longitude. As a result, the variance of the prediction is significantly decreased. (Right) Feature importance gained from a newly constructed BayesNAM for each location. The interaction term is highly important when distinguishing Yosemite from San Francisco and Los Angeles.}
    \label{fig:hist2}
\end{figure*}

In addition to data insufficiencies, a high level of inconsistency can reveal structural limitations within the model. In \Figref{fig:CA_Longitude}, we plot the results of NAMs and BayesNAM for longitude. These functions show higher housing prices in \texttt{San Francisco} (around -122.5) and \texttt{Los Angeles} (around -118.5), consistent with previous findings \cite{agarwal2021neural}. However, BayesNAM finds that there exist inconsistent explanations between these two cities, particularly within the longitude range of -120 to -119.

We hypothesize that this inconsistency is due to significant variations in housing prices (target variable) across different latitudes. \Figref{fig:CA_map} illustrates the distribution of housing prices in California, with red circles indicating higher prices and larger circles representing higher volumes of houses. As shown in \Figref{fig:CA_map}, while \texttt{Santa Barbara}, \texttt{Yosemite National Park}, and \texttt{Fresno} are on similar longitudes, \texttt{Santa Barbara} exhibits a substantial price gap compared to the others. Additionally, since \texttt{Yosemite National Park} and \texttt{Fresno} are near the same latitude as \texttt{San Francisco}, NAM might struggle to accurately predict housing prices without the interaction term between \texttt{Latitude} and \texttt{Longitude}.

Based on this observation, we train a NAM with an interaction term between \texttt{Latitude} and \texttt{Longitude}. This model achieved a much better performance (RMSE: $0.506 \pm 0.005$) than without the interaction term (RMSE: $0.556 \pm 0.009$). In addition, The significance of the interaction term is particularly evident near \texttt{Yosemite}. When BayesNAM is trained with this interaction term, it also shows reduced variance in the longitude range of -120 to -119. This suggests that the high variance can highlight potential structural limitations within models. Moreover, considering that existing methods such as NA$^2$M \cite{agarwal2021neural}, which incorporate all possible interaction terms, incur heavy computational costs and diminished explainability, BayesNAM offers a promising alternative by effectively selecting the most important interaction terms.

\section{Conclusion}
In this work, we identified and analyzed the inconsistent explanations of NAMs. We highlighted the importance of acknowledging these inconsistencies and introduced a new framework, BayesNAM, which leverages inconsistency to provide more reliable explanations. Through empirical validation, we demonstrated that BayesNAM effectively explores diverse explanations and provides external explanations such as insufficient data or model limitations within the data model. We hope our research contributes to the development of trustworthy models.


\bibliographystyle{IEEEtran}
\bibliography{IEEEabrv,ms} 

\newpage

\onecolumn

\section*{Supplements to \customtitle}
\subsection*{Proofs}

\subsubsection*{Proof for Lemma \ref{lem:lemma1}.}
\begin{proof}
Let $h(x_2, \cdots, x_d) = \sign(\sum_{i=2}^{d} x_i / (d-1))$. Then, the accuracy of $h(x_2, \cdots, x_d)$ becomes
\begin{align*}
    \mathbb{P}&[h(x_2, \cdots, x_d) = y]=\mathbb{P}\left[ \frac{y}{d-1}\sum_{i=2}^{d} \mathcal{N}(\lambda y, \sigma^2) > 0 \right]
    = \Phi_{X \sim \mathcal{N}\left(0, \sigma^2/(d-1)\right)}(\lambda).
\end{align*}
\end{proof}

\subsubsection*{Preliminary Lemma for Proof for \Thmref{thm:feature_dropout}.} 

\begin{lemma}
\label{lem:lem2}
Let $q_j=\Phi_{Z}(\lambda\sqrt{j}/\sigma)$. Then, the following inequality holds:
\begin{align*}
(q_{j+1}-q_j) \leq \frac{1}{\sqrt{2\pi}}\int_{\sqrt{j\text{ln}((j+1)/j)}}^{\sqrt{(j+1)\text{ln}((j+1)/j)}} e^{- \frac{1}{2}t^2}dt, 
\end{align*}
Moreover,
\begin{align*}
    (j+1)\frac{1}{\sqrt{2\pi}}\int_{\sqrt{j\text{ln}((j+1)/j)}}^{\sqrt{(j+1)\text{ln}((j+1)/j)}} e^{- \frac{1}{2}t^2}dt
\end{align*}
decreases with respect to $j$, so the upper bound for $j \geq 2$ is $\frac{3}{\sqrt{2\pi}}\int_{\sqrt{2\text{ln}(3/2)}}^{\sqrt{3\text{ln}(3/2)}} e^{- \frac{1}{2}t^2}dt.$
\end{lemma}

\begin{proof}
To find the upper bound of $q_{j+1}-q_j$, we take the derivative with respect to $a$ as follows:
\begin{align*}
\frac{\partial}{\partial a} (q_{j+1} - q_{j}) = \frac{1}{\sqrt{2\pi}}(\sqrt{j+1}e^{-\frac{j+1}{2}a^2} - \sqrt{j}e^{-\frac{j}{2}a^2}) = 0 \quad \text{at} \quad a = \sqrt{\text{ln}(j+1/j)}.
\end{align*}
Thus, the first inequality holds. Next, we take the derivative with respect to $j$ to verify that the left-hand side decreases.
\begin{align*}
    \frac{\partial}{\partial j} 
    (j+1)\int_{\sqrt{j\text{ln}((j+1)/j)}}^{\sqrt{(j+1)\text{ln}((j+1)/j)}}& e^{- \frac{1}{2}t^2}dt = \int_{\sqrt{j\text{ln}((j+1)/j)}}^{\sqrt{(j+1)\text{ln}((j+1)/j)}} e^{- \frac{1}{2}t^2}dt \\&+ (j+1)\left[\frac{( \frac{j}{j+1})^{\frac{j+1}{2}}}{2\sqrt{(j+1)\ln(j+1/j)}} \left\{ \ln\frac{j+1}{j} - \frac{1}{j} \right\} - \frac{( \frac{j}{j+1})^{\frac{j}{2}}  }{2\sqrt{j\ln(j+1/j)}} \left\{\ln\frac{j+1}{j}- \frac{1}{j+1} \right\}
    \right]    
\end{align*}
Simplifying this equation:
\begin{align*}
    (j+1)&\left[\frac{( \frac{j}{j+1})^{\frac{j+1}{2}}}{2\sqrt{(j+1)\ln(j+1/j)}} \left\{ \ln\frac{j+1}{j} - \frac{1}{j} \right\}- \frac{( \frac{j}{j+1})^{\frac{j}{2}}  }{2\sqrt{j\ln(j+1/j)}} \left\{\ln\frac{j+1}{j}- \frac{1}{j+1} \right\}
    \right] \\&=  \frac{j+1}{2\sqrt{\ln(j+1/j)}} \left( \frac{j}{j+1} \right)^{\frac{j}{2}}
    \times \left[\frac{\sqrt{j}}{{j+1}} (\ln\frac{j+1}{j} - \frac{1}{j}) - \frac{1}{\sqrt{j}} (\ln\frac{j+1}{j}- \frac{1}{j+1})    \right]\\&= -\frac{1}{2} \left( \frac{j}{j+1} \right)^{\frac{j}{2}}
    \sqrt{\ln\frac{j+1}{j}}\cdot\frac{1}{\sqrt{j}}
\end{align*}
Since
\begin{align*}
    \int_{\sqrt{j\text{ln}((j+1)/j)}}^{\sqrt{(j+1)\text{ln}((j+1)/j)}} e^{- \frac{1}{2}t^2}dt\leq \sqrt{\ln\frac{j+1}{j}} (\sqrt{j+1} - \sqrt{j})\left( \frac{j}{j+1} \right)^{\frac{j}{2}},
\end{align*}
we have
\begin{align*}
 \int_{\sqrt{j\text{ln}((j+1)/j)}}^{\sqrt{(j+1)\text{ln}((j+1)/j)}} e^{- \frac{1}{2}t^2}dt - \frac{1}{2} \left( \frac{j}{j+1} \right)^{\frac{j}{2}}
    \sqrt{\ln\frac{j+1}{j}}\cdot\frac{1}{\sqrt{j}} \leq \sqrt{\ln\frac{j+1}{j}} \left( \frac{j}{j+1} \right)^{\frac{j}{2}}(\sqrt{j+1} - \sqrt{j} - \frac{1}{2\sqrt{j}}) \leq 0,
\end{align*} 
for $j \geq 2$.
\end{proof}

\subsubsection*{Proof for \Thmref{thm:feature_dropout}.} 
\begin{proof}
Let $q_j=\Phi_{Z}(\lambda\sqrt{j}/\sigma)$. Then, $\mathcal{P}(k, \tau)$ can be formalized as follows:
\begin{equation}
    \mathcal{P}(k, \tau) = \sum^{k}_{j=1} q_j \binom kj (1-\tau)^j \tau^{k-j}.
\end{equation}

Additionally, let the difference between $\mathcal{P}(k, \tau)$ and $\mathcal{P}(1, \tau)$ as follows:
\begin{align*}
\Delta \mathcal{P}(k, \tau) := \mathcal{P}(k, \tau) - \mathcal{P}(1, \tau) &= \sum^{k}_{j=1} q_j \binom kj (1-\tau)^j \tau^{k-j} - q_1(1-\tau).
\end{align*}
It is trivial that $\Delta \mathcal{P}(k,0) = q_k-q_1 > 0$ $\forall k \geq 3$. 

To show that $\frac{\partial}{\partial \tau}\Delta \mathcal{P}(k,\tau)>0$ for $\tau\in[0, \frac{1}{2}]$, we first prove that $\frac{\partial}{\partial \tau}\Delta \mathcal{P}(3, \tau) > 0$ for $\tau\in[0, \frac{1}{2}]$. For $k=3$, we have
\begin{align*}
\Delta \mathcal{P}(3, \tau) &= \sum^{3}_{j=1} q_j \binom 3j (1-\tau)^j \tau^{3-j} - q_1(1-\tau)\\
&= (-3q_1 + 3q_2 -q_3)\tau^3 + (3q_1 - 6q_2 + 3q_3) \tau^2 + (q_1 + 3q_2 - 3q_3)\tau +(-q_1 + q_3),
\end{align*}
and
\begin{align}\label{eq:deltap_3}
    \frac{\partial}{\partial \tau}\Delta \mathcal{P}(3, \tau) &= (-9q_1 + 9q_2 - 3q_3)\tau^2 + (6q_1 -12q_2 +6q_3)\tau + (q_1 + 3q_2 - 3q_3).
\end{align}

For simplicity, let us denote $a=\lambda/\sigma$ so that $q_j=\Phi_{Z}(a\sqrt{j})$. Since $\frac{\partial}{\partial a} \Phi_{Z}(a\sqrt{j}) = \sqrt{j} \cdot \phi(a\sqrt{j})$,
\begin{align}\label{eq:temp_1}
\frac{\partial}{\partial a} (q_2 - q_1) = \frac{\partial}{\partial a} (\Phi_{Z}(\sqrt{2}a) - \Phi_{Z}(a)) =  \sqrt{2} \phi(a\sqrt{2}) + \phi(a),
\end{align}
where $\phi(x) = \frac{1}{\sqrt{2\pi}} e^{-\frac{1}{2}x^2}$. \eqref{eq:temp_1} becomes 0 when
\begin{align*}
\frac{\partial}{\partial a} (q_2 - q_1) = \frac{1}{\sqrt{2\pi}}(\sqrt{2}e^{-a^2} - e^{-\frac{1}{2}a^2}) = 0 \quad \text{at} \quad a = \sqrt{\text{ln}2}.
\end{align*}

Since $q_j > \frac{1}{2}$ for $a > 0$ and $j > 0$, we have
\begin{align*}
q_2-q_1 \leq \frac{1}{\sqrt{2\pi}} \int_{\sqrt{\text{ln}2}}^{ \sqrt{2\text{ln}2}} e^{- \frac{1}{2}t^2}dt \approx 0.08302 < \frac{1}{6} < \frac{1}{3} q_3,
\end{align*}
so that $-9q_1 +9q_2-3q_3 < 0$ for any $a$.

Similarly, we can easily show that 
\begin{align} \label{eq:bd}
\frac{\partial}{\partial a} (q_1 + 3q_2 - 3q_3) = \frac{1}{\sqrt{2\pi}} (e^{-\frac{1}{2}a^2} + 3\sqrt{2}e^{-a^2} - 3\sqrt{3}e^{-\frac{3}{2}a^2}) = \frac{1}{\sqrt{2\pi}}(x+3\sqrt{2}x^2 - 3\sqrt{3}x^3),
\end{align}
where $x = e^{-\frac{1}{2}a^2} \in (0, 1]$ for $a^2 \in [0, \infty)$.
The solutions that make \eqref{eq:bd} equal to zero can be explicitly calculated as $x = 0, \frac{\sqrt{6} - \sqrt{6+4\sqrt{3}}}{6}<0$ and $\frac{\sqrt{6} + \sqrt{6+4\sqrt{3}}}{6} > 1$. The minimum value of $q_1 + 3q_2 -3q_3$ is attained when $a$ lies on the boundary.
Thus, $\frac{\partial}{\partial \tau}\Delta \mathcal{P}(3, \tau)|_{\tau=0} \geq 0.5$.

We will now demonstrate that $\frac{\partial}{\partial \tau}\Delta \mathcal{P}(3, \tau)|_{\tau=\frac{3}{8}}>0$ by showing that $4*\frac{\partial}{\partial \tau}\Delta \mathcal{P}(3, \tau)|_{\tau=\frac{1}{2}}=7q_1 - 3q_2 -3q_3>0$.
Note that  
\begin{equation}
    7q_1 - 3q_2 -3q_3 = \frac{7}{\sqrt{2\pi}}\int _{-\infty}^{a}e^{-\frac{1}{2}t^2}dt - \frac{3}{\sqrt{2\pi}}\int _{-\infty}^{a\sqrt{2}}e^{-\frac{1}{2}t^2}dt - \frac{3}{\sqrt{2\pi}}\int _{-\infty}^{a\sqrt{3}}e^{-\frac{1}{2}t^2}dt.
\end{equation}
By taking the derivative with respect to $a$, we have 
\begin{equation}\label{eq:tem}
    \frac{\partial}{\partial a}(7q_1-3q_2-3q_3) = \frac{1}{\sqrt{2\pi}} (7e^{-\frac{1}{2}a^2} -3\sqrt{2}e^{-a^2} - 3\sqrt{3}e^{-\frac{3}{2}a^2}) = \frac{1}{\sqrt{2\pi}}(7x-3\sqrt{2}x^2 - 3\sqrt{3}x^3),
\end{equation}
where $x = e^{-\frac{1}{2}a^2}$. The solutions that make \eqref{eq:tem} equal to zero can be explicitly calculated as $x = 0, \frac{\pm \sqrt{6+28\sqrt{3}}-\sqrt{6}}{6}$. Since $e^{-\frac{1}{2}a^2}\in(0,1]$ and strictly decreases, the minimum occurs at $e^{-\frac{1}{2}\tilde{a}^2} = \frac{\sqrt{6+28\sqrt{3}}-\sqrt{6}}{6}$, where $\tilde{a} = \sqrt{2\ln{\frac{6}{\sqrt{6+28\sqrt{3}}-\sqrt{6}}}}$. 
Thus, we obtain the following inequality:
\begin{align*}
7q_1-3q_2-3q_3  &\geq  \frac{7}{\sqrt{2\pi}}\int _{-\infty}^{\tilde a}e^{-\frac{1}{2}t^2}dt - \frac{3}{\sqrt{2\pi}}\int _{-\infty}^{\tilde a\sqrt{2}}e^{-\frac{1}{2}t^2}dt - \frac{3}{\sqrt{2\pi}}\int _{-\infty}^{\tilde a\sqrt{3}}e^{-\frac{1}{2}t^2}dt
> 0.1217 > 0.
\end{align*}
Thus implies that $\frac{\partial}{\partial \tau}\Delta \mathcal{P}(3, \tau)\lvert_{\tau=\frac{1}{2}} > 0.0304 >0$. Since $\frac{\partial}{\partial \tau}\Delta \mathcal{P}(3, \tau)$ with a negative leading coefficient, it follows that $\frac{\partial}{\partial \tau}\Delta \mathcal{P}(3, \tau) >0$ for $\tau\in[0, \frac{1}{2}]$.

Next, we demonstrate that $\Delta \mathcal{P}(k, \tau)$ increases as $\tau$ increases for $\tau\in[0, \frac{1}{2}]$ and $k\geq 3$. To begin, we decompose $\Delta \mathcal{P}(k+1, \tau)$ as follows:
\begin{align*}
\Delta \mathcal{P}(k+1, \tau) &= \sum^{k+1}_{j=1} q_j \binom {k+1}j (1-\tau)^j \tau^{k+1-j} - q_1(1-\tau)\\ 
&= \sum^{k}_{j=1} q_j \left( \binom k j +\binom k {j-1} \right) (1-\tau)^j \tau^{k+1-j} + q_{k+1}(1-\tau)^{k+1} - q_1(1-\tau)\\
&= \Delta \mathcal{P}(k,\tau) + \sum^{k}_{j=1} ( q_{j+1} -q_j ) \binom k j (1-\tau)^{j+1} \tau^{k-j} + q_1(1-\tau)\tau^k\\
&= \Delta \mathcal{P}(3,\tau) + \sum^{k}_{l=3} \sum^{l}_{j=1} ( q_{j+1} -q_j ) \binom l j (1-\tau)^{j+1} \tau^{l-j} + q_1(1-\tau)\tau^l\\
&= \Delta \mathcal{P}(3,\tau) + q_1 (1-\tau) \sum^{k}_{l=3} \tau^l + (q_2-q_1)(1-\tau)^2 \sum^{k}_{l=3}\binom l 1 \tau^{l-1} + (q_3-q_2)(1-\tau)^3 \sum^{k}_{l=3}\binom l 2 \tau^{l-2}\\
&  \quad + \sum^{k}_{j=3} ( q_{j+1} -q_j )(1-\tau)^{j+1} \sum^{k}_{l=j}  \binom l j  \tau^{l-j}.
\end{align*}

By taking the derivative with respect to $\tau$, we have
\begin{align*}
\frac{\partial}{\partial \tau}\Delta \mathcal{P}(k + 1,\tau) &=\frac{\partial}{\partial \tau}\Delta \mathcal{P}(3,\tau) + q_1 \sum^{k}_{l=3} \tau^{l-1} \Big\{ \binom l 1 - \binom {l+1} 1 \tau \Big\} + 2(q_2-q_1)(1-\tau) \sum^{k}_{l=3} \tau^{l-2} \Big\{ \binom l 2 - \binom {l+1} 2 \tau \Big\} \\
& \quad + 3(q_3-q_2)(1-\tau)^2 \sum^{k}_{l=3} \tau^{l-3} \Big\{ \binom l 3 - \binom {l+1} 3 \tau \Big\} \\ 
& \quad + \sum^{k}_{j=3} (j+1)(q_{j+1}-q_j)(1-\tau)^j \Big[- 1 + \sum^{k}_{l=j+1} \tau^{l-j-1} \Big\{ \binom l {j+1} - \binom {l+1} {j+1} \tau \Big\}\Big]  \\
&= \frac{\partial}{\partial \tau}\Delta \mathcal{P}(3,\tau) + q_1(3\tau^2 - \binom {k+1} 1 \tau^k) + 2(q_2-q_1)(1-\tau)(\binom 3 2 \tau - \binom {k+1} 2 \tau^{k-1})\\
& \quad + 3(q_3-q_2)(1-\tau)^2(1 - \binom {k+1} 3 \tau^{k-2}) - \sum^{k}_{j=3} (j+1)(q_{j+1}-q_j)\binom {k+1} {j+1} (1-\tau)^j \tau^{k-j}\\
&= \frac{\partial}{\partial \tau}\Delta \mathcal{P}(3,\tau) + 3 q_1 \tau^2 + 6(q_2-q_1)\tau(1-\tau) +3(q_3-q_2)(1-\tau)^2\\
& \quad - (k+1) \sum^{k}_{j=0} (q_{j+1}-q_j)\binom k j (1-\tau)^j \tau^{k-j} - (k+1)q_0\tau^k \\
&= \frac{\partial}{\partial \tau}\Delta \mathcal{P}(3,\tau) + q_1 \tau^2 (3- (k+1)\tau^{k-2}) + (q_2-q_1)\tau(1-\tau)(6-(k+1)k\tau^{k-2}) \\ & \quad +3 (q_3-q_2)(1-\tau)^2- \sum^{k}_{j=2} (j+1)(q_{j+1}-q_j)\binom {k+1} {j+1} (1-\tau)^j \tau^{k-j}
\end{align*}

By Lemma \ref{lem:lem2}, for $j \geq 2$,
\begin{align*}
(j+1)(q_{j+1}-q_j) \leq (j+1)\frac{1}{\sqrt{2\pi}}\int_{\sqrt{j\text{ln}((j+1)/j)}}^{\sqrt{(j+1)\text{ln}((j+1)/j)}} e^{- \frac{1}{2}t^2}dt \leq \frac{3}{\sqrt{2\pi}}\int_{\sqrt{2\text{ln}(3/2)}}^{\sqrt{3\text{ln}(3/2)}} e^{- \frac{1}{2}t^2}dt \leq 0.147.   
\end{align*}

Additionally, since $\frac{\partial}{\partial \tau}\Delta \mathcal{P}(3,\tau)$ is quadratic function, we have

\begin{align*}
\frac{\partial}{\partial \tau}\Delta \mathcal{P}(3,\tau) \geq  (1-2\tau) \frac{\partial}{\partial \tau}\Delta \mathcal{P}(3,\tau)|_{\tau=0} +2\tau \frac{\partial}{\partial \tau}\Delta \mathcal{P}(3,\tau)|_{\tau=\frac{1}{2}} \geq 0.5 - 0.94 \tau.
\end{align*}

Moreover, since $(k+1)\tau^{k-2}$ decreases for $k\geq3$, we have 
\begin{align*}
    \frac{\partial}{\partial \tau}\Delta \mathcal{P}(k + 1,\tau) &\geq \frac{\partial}{\partial \tau}\Delta \mathcal{P}(3,\tau) + q_1 \tau^2 (3- (k+1)\tau^{k-2})- \sum^{k}_{j=2} (j+1)(q_{j+1}-q_j)\binom {k+1} {j+1} (1-\tau)^j \tau^{k-j} \\ &\geq 0.5 - 0.94 \tau + \frac{1}{2}\tau^2(3-4\tau) - \frac{0.147}{1-\tau}. 
\end{align*}

Therefore, we proved that $0.5 - 0.94 \tau + \frac{1}{2}\tau^2(3-4\tau) - \frac{0.147}{1-\tau} > 0 $ for $\tau \in [0, \frac{3}{8}]$, $\frac{\partial}{\partial \tau}\Delta \mathcal{P}(k + 1,\tau) \geq 0$ for $\tau \in [0, \frac{3}{8}]$ for $k\geq 3$.

Now, we will show that $\frac{\partial}{\partial \tau}\Delta \mathcal{P}(4,\tau) \geq 0$ and  $\frac{\partial}{\partial \tau}\Delta \mathcal{P}(5,\tau) \geq 0$ for $\tau \in [\frac{3}{8}, \frac{1}{2}]$.
First,
\begin{align*}
\frac{\partial}{\partial \tau}\Delta \mathcal{P}(4,\tau) &= \frac{\partial}{\partial \tau}\Delta \mathcal{P}(3,\tau) + 3 q_1 \tau^2 + 6(q_2-q_1)\tau(1-\tau) +3(q_3-q_2)(1-\tau)^2  - 4\sum^{3}_{j=0} (q_{j+1}-q_j)\binom 3 j (1-\tau)^j \tau^{3-j} - 4q_0\tau^3 \\
&= \frac{\partial}{\partial \tau}\Delta \mathcal{P}(3,\tau) + q_1\tau^2(3-4\tau) + 6(q_2-q_1)\tau(1-\tau)(1-2\tau)+3(q_3-q_2)(1-\tau)^2(1-4\tau) - 4(q_4-q_3)(1-\tau)^3.
\end{align*}

By using Lemma \ref{lem:lem2}, we can have the following inequalities:
\begin{align*}
q_3 - q_2 \leq \frac{1}{\sqrt{2\pi}}\int_{\sqrt{2\text{ln}(3/2)}}^{\sqrt{3\text{ln}(3/2)}} e^{- \frac{1}{2}t^2}dt \leq 0.049,
\end{align*}
and
\begin{align*}
q_4 - q_3 \leq \frac{1}{\sqrt{2\pi}}\int_{\sqrt{3\text{ln}(4/3)}}^{\sqrt{4\text{ln}(4/3)}} e^{- \frac{1}{2}t^2}dt \leq 0.035.
\end{align*}

Since $q_3 - q_2 \geq q_4 - q_3$, we can also the following inequalities:
\begin{align*}
 3(q_3-q_2)(1-\tau)^2(4\tau-1) + 4(q_4-q_3)(1-\tau)^3 \leq (q_3-q_2)(1-\tau)^2(1+8\tau) \leq \frac{25}{16}(q_3-q_2) \leq 0.077,
\end{align*}
and
\begin{align*}
\frac{\partial}{\partial \tau}\Delta \mathcal{P}(3,\tau)+q_1 \tau^2(3-4\tau) \geq \frac{\partial}{\partial \tau}\Delta \mathcal{P}(3,\tau)|_{\tau=\frac{1}{2}} + \frac{27}{128} q_1 \geq 0.03 + 0.1054 = 0.1354.
\end{align*}
Therefore, $\frac{\partial}{\partial \tau}\Delta \mathcal{P}(4,\tau) \geq \frac{\partial}{\partial \tau}\Delta \mathcal{P}(3,\tau)+q_1 \tau^2(3-4\tau) -(3(q_3-q_2)(1-\tau)^2(4\tau-1)+ 4(q_4-q_3)(1-\tau)^3) \geq 0.1354 - 0.077 > 0$.

Similarly, we have
\begin{align*}
\frac{\partial}{\partial \tau}\Delta \mathcal{P}(5,\tau) &= \frac{\partial}{\partial \tau}\Delta \mathcal{P}(3,\tau) + 3 q_1 \tau^2 + 6(q_2-q_1)\tau(1-\tau) +3(q_3-q_2)(1-\tau)^2  \\
& \quad - 5\sum^{4}_{j=0} (q_{j+1}-q_j)\binom 4 j (1-\tau)^j \tau^{4-j} - 5q_0\tau^4 \\
&= \frac{\partial}{\partial \tau}\Delta \mathcal{P}(3,\tau) + q_1\tau^2(3-5\tau^2) + (q_2-q_1)\tau(1-\tau)(6-20\tau^2)+3(q_3-q_2)(1-\tau)^2(1-10\tau^2) \\
& \quad - 20(q_4-q_3)(1-\tau)^3\tau - 5(q_5-q_4)(1-\tau)^4 \\
& \geq \frac{\partial}{\partial \tau}\Delta \mathcal{P}(3,\tau) + q_1\tau^2(3-5\tau^2)-(q_3-q_2)(1-\tau)^2(15\tau^2+10\tau+2) \\
&\geq 0.03 + \frac{1323}{4096}q_1 - \frac{12575}{4096}(q_3-q_2) \geq 0.03 + 0.1623 - 0.1505 > 0.
\end{align*}

For $k \geq 6$, we can easily show that
\begin{align*}
    \frac{\partial}{\partial \tau}\Delta \mathcal{P}(k,\tau) &\geq \frac{\partial}{\partial \tau}\Delta \mathcal{P}(3,\tau) + q_1 \tau^2 (3- k\tau^{k-3})- \sum^{k-1}_{j=2} (j+1)(q_{j+1}-q_j)\binom {k} {j+1} (1-\tau)^j \tau^{k-1-j} \\ &\geq 0.5 - 0.94 \tau + \frac{1}{2}\tau^2(3-6\tau^3) - \frac{0.147}{1-\tau}. 
\end{align*}

Since $0.5 - 0.94 \tau + \frac{1}{2}\tau^2(3-6\tau^3) - \frac{0.147}{1-\tau} > 0 $ for $\tau \in [\frac{3}{8},\frac{1}{2}]$, the inequality holds.

Therefore, $\frac{\partial}{\partial \tau}\Delta \mathcal{P}(k,\tau) \geq 0$ for $\tau \in [0, \frac{1}{2}]$ for $k \geq 3$.
\end{proof}

\subsection*{Training Details}

In \Secref{sec:exp}, we conducted experiments on five different datasets: Credit Fraud \cite{dal2015adaptive}, FICO \cite{fico2018fico}, COMPAS \cite{compas2016compas}, California Housing (CA Housing) \cite{pace1997sparse}, and Boston \cite{harrison1978hedonic}. Here, we provide a summary of the characteristics in \Tabref{tab:dataset} and detailed explanations to ease the understanding of the experimental interpretation.

\begin{table}[h!]
\caption{List of datasets and their characteristics.}
\label{tab:dataset}
\centering
\begin{tabular}{lrrrr}
\hline
\textbf{Data}&\textbf{\# Train} &\textbf{\# Test }&\textbf{\# Features }& \textbf{Task type}    \\ \hline 
Credit Fraud       & 227,845  & 56,962  & 30         & Classification \\ 
FICO               & 8,367    & 2,092   & 23         & Classification \\ 
COMPAS             & 13,315   & 3,329   & 17         & Classification \\ 
CA Housing & 16,512   & 4,128   & 8          & Regression     \\
Boston    & 404     & 102    & 13         & Regression     \\ 
\hline
\end{tabular}
\end{table}
 
\begin{itemize}
    \item \textbf{Credit Fraud:} This dataset focuses on predicting fraudulent credit card transactions and is highly imbalanced. Due to confidentiality concerns, the features are represented as principal components obtained through PCA. 
    \item \textbf{FICO:} This dataset aims to predict the risk performance of consumers, categorizing them as either ``Bad'' or ``Good'' based on their credit.
    \item \textbf{COMPAS:} This dataset aims to predict recidivism, determining whether an individual will reoffend or not.
    \item \textbf{CA Housing:} This dataset aims to predict the median house value for districts in California based on data derived from the 1990 U.S. Census.
    \item \textbf{Boston:} This dataset focuses on predicting the median value of owner-occupied homes in the Boston area, using data collected by the U.S. Census Service.
\end{itemize}

For each dataset, we conducted experiments using five different random seeds. To ensure a fair comparison with the prior work \cite{agarwal2021neural}, we adopted 5-fold cross-validation for datasets that train-test split was not provided.

\begin{table}[h!]
\caption{Comparison between the structures of previous NAM \cite{agarwal2021neural} and the proposed.}
\label{tab:resnet}
\centering
\begin{tabular}{c|cc|cc}
\hline
\textbf{Data}&\textbf{NAM in \cite{agarwal2021neural}} &\textbf{AUC($\uparrow$)}&\textbf{NAM in ours}& \textbf{AUC($\uparrow$)}    \\ \hline 
Credit Fraud       & ExU+ReLU-$1$  & 0.980$\pm$0.00  & ResBlocks+ReLU         & \textbf{0.990$\pm$0.00} \\ 
COMPAS             & Linear+ReLU   & 0.737$\pm$0.01   & ResBlocks+ReLU         &  \textbf{0.771$\pm$0.05} \\
\hline
\end{tabular}
\end{table}

In the original paper of NAM \cite{agarwal2021neural}, the authors proposed exp-centered (ExU) units, which can be formalized as follows:
\begin{equation}
    \text{ExU}(x) = h(e^w (x-b)),
\end{equation}
where $w$ and $b$ are weight and bias parameters, and $h(\cdot)$ is an activation function. ExU units were proposed to model jagged functions, enhancing the expressiveness of NAM. The authors also explored the use of ReLU-$n$ activation that bounds the ReLU activation at $n$ and they found it to be beneficial for specific datasets. However, we discover that their convergence is relatively unstable compared to fully connected networks.

\begin{figure}[ht!]
    \centering
    \includegraphics[width=0.6\linewidth]{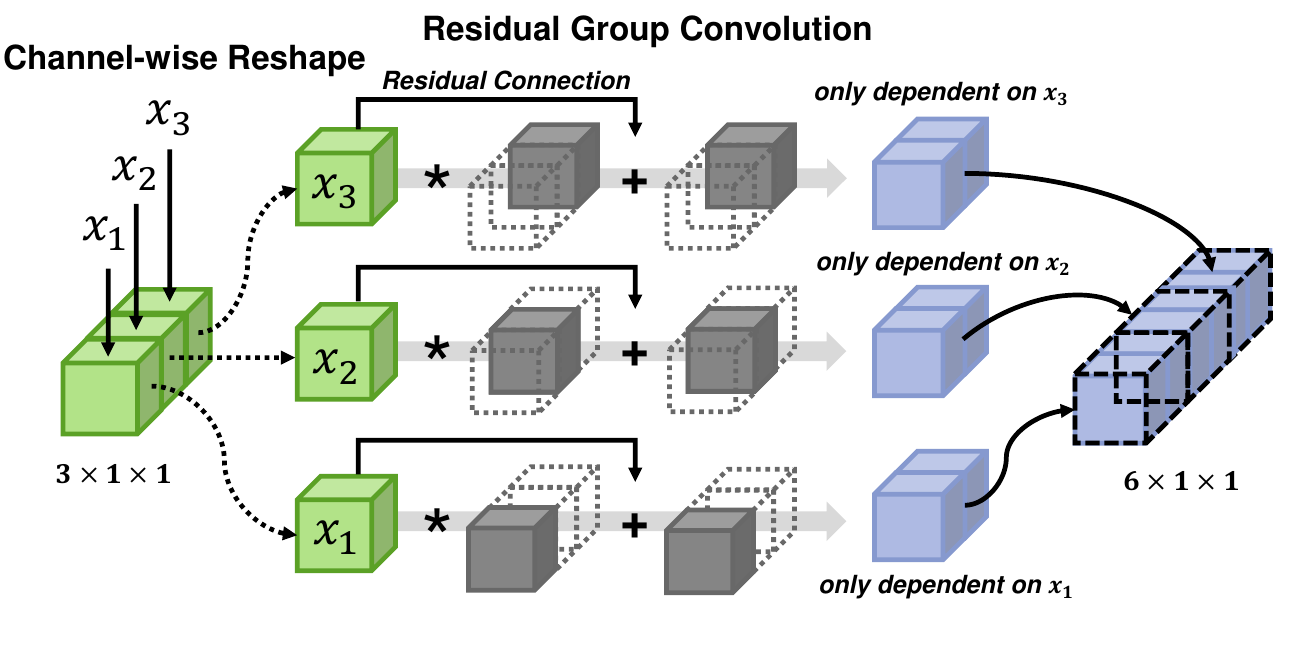}
    \caption{Illustration of the use of group convolution and its effect on reducing computational costs.}
    \label{fig:gc}
\end{figure}

To address this issue, we employ ResNet blocks in our approach. Specifically, we utilize a ResNet block with group convolution layers, where each layer is followed by BatchNorm and ReLU activation. This modification enables stable training across various datasets and further significantly improves performance. Following the standard ResNet structure, our model includes one input layer with BatchNorm, ReLU, and Dropout, three ResNet blocks, and one output layer. We find that a dimension of $32$ for each layer is sufficient to achieve superior performance. In \Tabref{tab:resnet}, we compare the structures and performances of NAM in \cite{agarwal2021neural} with our proposed structure for common datasets under the same setting in \cite{agarwal2021neural}. Our proposed structure shows improved performance while requiring lower computational costs.
By combining the group convolution integration proposed in 
\cite{radenovic2022neural}, the basic structure of our neural additive model becomes \Figref{fig:gc}.
Given the feature dimension $d=3$, the inputs are reshaped accordingly within the channel dimension. Subsequently, we employ group convolution using $d$ groups, where each kernel is individually applied to its corresponding channel. Then, residual connections are employed for each function $f_i$.

\begin{table}[ht!]
\caption{Selected hyper-parameters by grid-search.}
\label{tab:grid_search}
\centering
\begin{tabular}{l|ccccc}
\hline
\textbf{Params}&\textbf{COMPAS} &\textbf{Credit}&\textbf{FICO}& \textbf{Boston}&\textbf{CA Housing}    \\ \hline
Learning rate $\eta$ & 0.01 & 0.01 & 0.01 & 0.001 & 0.01 \\ 
Dropout rate $\psi$ & 0.1 & 0.3 & 0.0 & 0.0 & 0.0 \\
Batch-size $B$ & 1,024 & 1,024 & 2,048 & 128 & 2,048 \\ 
Feature Dropout $\tau$ & 0.2 & 0.1 & 0.4 & 0.1 & 0.1 \\
\hline
\end{tabular}
\end{table}

We performed grid searches to determine the best settings for achieving high performance. First, for NAM, we explored the learning rate $\eta\in[0.1, 0.01, 0.001, 0.0001]$, the dropout rate in the input layer $\psi \in [0.0, 0.1, 0.2, 0.3, 0.4, 0.5]$, and the batch size $B\in[128, 256, 512, 1024, 2048]$. We then explored the additional hyper-parameters for BayesNAM while fixing other variables the same as NAM. Following \cite{liu2018adv}, we searched for the initial standard deviation vector $s_0$, and found that $s_0=1e^{-4}$ provided the most stable performance. We explored the feature dropout probability $\tau\in[0.1, 0.2, 0.3, 0.4, 0.5]$. In all experiments, we used SGD with cosine learning rate decay, a momentum of 0.9, and weight decay of $5e^{-4}$ over 100 epochs. The selected hyper-parameter settings are provided in \Tabref{tab:grid_search}.

\subsection*{Ablation Study}

\begin{figure}[ht!]
    \centering
    \includegraphics[height=0.25\linewidth]{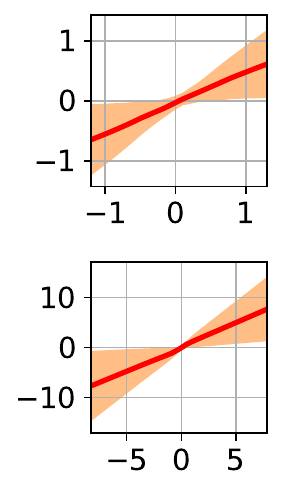}
    \includegraphics[height=0.25\linewidth]{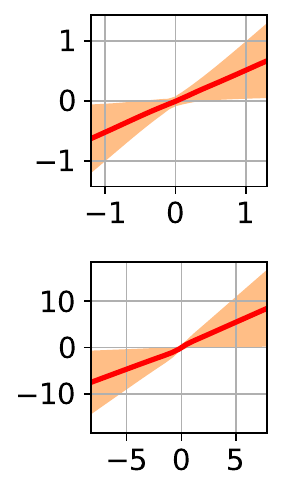}
    \includegraphics[height=0.25\linewidth]{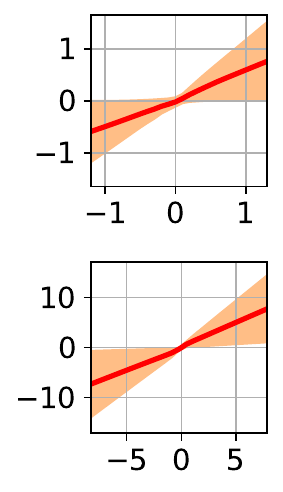}
    \includegraphics[height=0.25\linewidth]{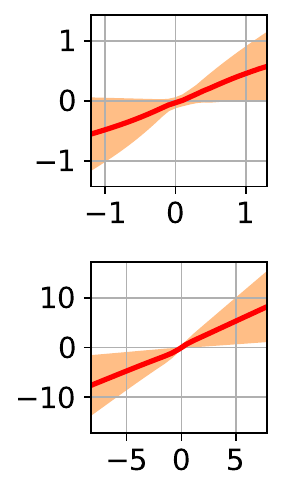}
    \includegraphics[height=0.25\linewidth]{figs/bayes_fd_small5.pdf}
    \caption{Consistency of BayesNAM for five different random seeds. Mapping functions of $f_1(x_1)$ (up row) and $f_2(x_2)$ (bottom row) obtained from BayesNAM with $\tau=0.1$ for five distinct random seeds. The same setting is used as in \Figref{fig:lam3}. The average function is plotted in red, with the min-max range indicated in orange. Each column corresponds to a different random seed. In contrast to NAM in \Figref{fig:lam3}, BayesNAM consistently exhibits similar mapping function distribution.}
    \label{fig:hist3}
\end{figure}

\begin{table*}[ht!]
\caption{Performance comparison between models on 5 different random seeds. Higher AUC is better ($\uparrow$) and lower RMSE is better ($\downarrow$).}
\label{tab:ablation}
\centering
\resizebox{0.8\linewidth}{!}{%
\begin{tabular}{|c|c|c|c|c|c|}
\hline
Model              & \begin{tabular}[c]{@{}c@{}}COMPAS\\ (AUC$\uparrow$)\end{tabular} & \begin{tabular}[c]{@{}c@{}}Credit\\ (AUC$\uparrow$)\end{tabular} & \begin{tabular}[c]{@{}c@{}}FICO\\ (AUC$\uparrow$)\end{tabular} & \begin{tabular}[c]{@{}c@{}}Boston   \\ (RMSE$\downarrow$)\end{tabular} & \begin{tabular}[c]{@{}c@{}}CA   Housing\\ (RMSE$\downarrow$)\end{tabular} \\ \hline

w/o FD          & 0.782$\pm$0.006   & 0.990$\pm$0.005   & 0.805$\pm$0.003   & 3.618$\pm$0.007   & 0.554$\pm$0.010   \\ 
w/ FD         & 0.784$\pm$0.009   & 0.991$\pm$0.003   & 0.804$\pm$0.001   & 3.620$\pm$0.011   & 0.556$\pm$0.007   \\ 
\hline
\end{tabular}%
}
\end{table*}

\Figref{fig:hist3} and \Tabref{tab:ablation} show the ablation study on the feature dropout. \Figref{fig:hist3} shows the consistent explanations of BayesNAM under the setting of \Figref{fig:lam3}. Compared to the results of NAM in \Figref{fig:lam3}, BayesNAM effectively explores diverse explanations, even across different random seeds, yielding more consistent results. In  \Tabref{tab:ablation}, we compare the performance of BayesNAM without feature dropout (denoted as 'naive Bayesian' in \Figref{fig:fd}) and with feature dropout. While feature dropout encourages greater exploration of diverse explanations, it does not always lead to improved performance across all datasets. Specifically, for regression tasks, feature dropout often results in performance degradation. This is left as future work, aiming to develop a framework that enhances both performance and explainability across all possible tasks and datasets.

\end{document}